\documentclass{article} 
\usepackage{iclr2027_conference,times}

\usepackage{amsmath,amsfonts,bm}

\def\eqref#1{equation~\ref{#1}}

\def\1{\bm{1}}

\def\ra{{\textnormal{a}}}

\def\rx{{\textnormal{x}}}

\def\rva{{\mathbf{a}}}

\def\erva{{\textnormal{a}}}

\def\ervx{{\textnormal{x}}}

\def\rmA{{\mathbf{A}}}

\def\vmu{{\bm{\mu}}}
\def\vtheta{{\bm{\theta}}}
\def\va{{\bm{a}}}

\def\ve{{\bm{e}}}

\def\vx{{\bm{x}}}

\def\eva{{a}}

\def\mA{{\bm{A}}}

\def\mH{{\bm{H}}}
\def\mI{{\bm{I}}}
\def\mJ{{\bm{J}}}

\def\mX{{\bm{X}}}

\def\mSigma{{\bm{\Sigma}}}

\DeclareMathAlphabet{\mathsfit}{\encodingdefault}{\sfdefault}{m}{sl}
\SetMathAlphabet{\mathsfit}{bold}{\encodingdefault}{\sfdefault}{bx}{n}
\newcommand{\tens}[1]{\bm{\mathsfit{#1}}}
\def\tA{{\tens{A}}}

\def\tX{{\tens{X}}}

\def\gG{{\mathcal{G}}}

\def\sA{{\mathbb{A}}}
\def\sB{{\mathbb{B}}}

\def\sS{{\mathbb{S}}}

\def\emA{{A}}

\newcommand{\etens}[1]{\mathsfit{#1}}

\def\etA{{\etens{A}}}

\newcommand{\E}{\mathbb{E}}

\newcommand{\R}{\mathbb{R}}

\newcommand{\KL}{D_{\mathrm{KL}}}
\newcommand{\Var}{\mathrm{Var}}

\newcommand{\Cov}{\mathrm{Cov}}

\newcommand{\normltwo}{L^2}
\newcommand{\normlp}{L^p}

\newcommand{\parents}{Pa} 

\usepackage{url}

\usepackage[utf8]{inputenc} 
\usepackage[T1]{fontenc}    
\usepackage{url}            
\usepackage{booktabs}       
\usepackage{amsfonts}       
\usepackage{nicefrac}       
\usepackage{microtype}      
\usepackage{xcolor}         
\usepackage{subfigure}
\usepackage{pgfplots}
\usepackage{pgfplotstable}
\usepackage{tikz}
\usepgfplotslibrary{groupplots}

\usepackage{graphicx}
\usepackage[font=small]{caption}

\usepackage{makecell}

\definecolor{cb_orange}{RGB}{213,94,0}
\definecolor{cb_green}{RGB}{34,136,51}
\definecolor{cbgreen}{RGB}{34,136,51}
\definecolor{sky_blue}{RGB}{204, 238, 255}
\definecolor{cb_purple}{RGB}{170, 51, 119}
\definecolor{cb_red}{RGB}{204, 51, 17}
\definecolor{cb_blue}{RGB}{0, 119, 187}
\definecolor{mydarkblue}{rgb}{0,0.08,0.45}
\definecolor{forestgreen}{RGB}{34,139,34}
\definecolor{periwinkle}{rgb}{0.4, 0.4, 0.8}
\definecolor{royalazure}{rgb}{0.0, 0.22, 0.66}
\definecolor{royalblue}{rgb}{0.0, 0.14, 0.4}
\definecolor{magenta}{RGB}{255, 0, 255}
\definecolor{myorange}{RGB}{191, 128, 64}
\definecolor{richlilac}{rgb}{0.71, 0.4, 0.82}
\definecolor{mydeeppink}{RGB}{255, 20, 147}
\usepackage[colorlinks=true,citecolor=mydeeppink,linkcolor=mydeeppink,urlcolor=royalazure]{hyperref}

\usepackage{listings}       
\lstdefinestyle{promptstyle}{basicstyle=\ttfamily\scriptsize, breaklines=true,
  breakindent=0pt, breakatwhitespace=false, postbreak={}, columns=fullflexible,
  keepspaces=true, showstringspaces=false, xleftmargin=0pt, frame=none, aboveskip=6pt,
  belowskip=6pt, upquote=true}
\usepackage{tcolorbox}
\tcbuselibrary{breakable}

\usepackage{amsmath}
\usepackage{amssymb}
\usepackage{mathtools}
\usepackage{amsthm}

\usepackage[capitalize,noabbrev]{cleveref}

\usepackage{wrapfig}
\usepackage{enumitem}
\usepackage{multirow}
\usepackage{algorithm}
\usepackage{algorithmic}

\theoremstyle{plain}

\theoremstyle{definition}

\theoremstyle{remark}

\definecolor{cbgreen}{RGB}{34,136,51}
\definecolor{cbblue}{RGB}{0, 119, 187}
\definecolor{cbred}{RGB}{204, 51, 17}
\definecolor{mygray}{RGB}{100, 100, 100}
\newcommand{\pinkcell}[1]{\cellcolor{teal!15}#1}
\newcommand{\planned}[1]{\textcolor{red}{#1}}
\newcommand{\placeholderfig}[2]{\IfFileExists{#1}{\includegraphics[width=#2]{#1}}{\fbox{\parbox[c][0.22\linewidth][c]{#2}{\centering\planned{Placeholder: #1}}}}}
\usepackage{bbding}          
\usepackage{pifont}          
\newcommand{\yes}{\textcolor{cb_green}{\ding{51}}}
\newcommand{\no}{\textcolor{cb_red}{\ding{55}}}
\usepackage{colortbl}        

\title{AutoRef: Harness Optimization for\\Agentic Multi-Reference Image Generation}

\author{%
  \textbf{Yuta~Oshima$^{1,*}$\quad
  Ku~Onoda$^{1,*}$\quad
  Yusuke~Iwasawa$^{1}$\quad
  Masahiro~Suzuki$^{1}$} \\
  \textbf{Yutaka~Matsuo$^{1}$\quad
  Hiroki~Furuta} \\
  \textsuperscript{1}The University of Tokyo \\
  \texttt{\{yuta.oshima, ku.onoda\}@weblab.t.u-tokyo.ac.jp}
}

\iclrfinalcopy 
\begin{document}

\maketitle

\begin{abstract}

Recent image generation models can take multiple reference images as input and combine them into a new image. 
However, multi-reference image generation remains challenging: models may omit or duplicate subjects from the references, or produce images in which multiple subjects appear unnaturally copied and pasted. 
Recent work has proposed image generation agents that combine image generation models, reasoning models, and a harness, which is an executable program that specifies how reference images are interpreted, how generation is performed, how outputs are diagnosed, and how the final image is selected. 
In multi-reference generation, however, references play different roles and outputs must satisfy many criteria at once, such as fidelity to each reference and the naturalness of the whole image, so many parts of the harness could be improved, from how references are processed to how outputs are diagnosed.
This makes it hard to predict which changes will improve performance and by how much, and good harnesses difficult to design by hand; indeed, human-written harnesses vary widely in performance.
We therefore propose \textbf{AutoRef}, which optimizes the harness automatically while keeping both models frozen: a coding agent iteratively rewrites the harness code.
AutoRef separates the tasks whose feedback informs proposals from the tasks used to select candidates, and continues the search from a beam of the top-ranked harnesses on the selection tasks. 
Using this procedure, we discover \textbf{AutoRef-Harness}, which improves the open-weight FLUX.2 [klein] 4B from 5.72 to 7.37 (out of 10) on held-out four-reference tasks of the MultiBanana benchmark, matching or exceeding proprietary models including Nano Banana Pro and GPT-Image-1.5. 
Without re-optimization, the same harness also improves results when the generator, number of references, benchmark, evaluator, or reasoning model differs from those used in the search. 
We release our code and AutoRef-Harness at \url{https://github.com/KuOnoda/AutoRef}.

\end{abstract}

\begingroup
\renewcommand\thefootnote{*}
\footnotetext{Equal contribution}
\endgroup

\begin{figure}[ht]
    \centering
    \begin{minipage}{0.33\linewidth}
        \centering
        \includegraphics[width=\linewidth]{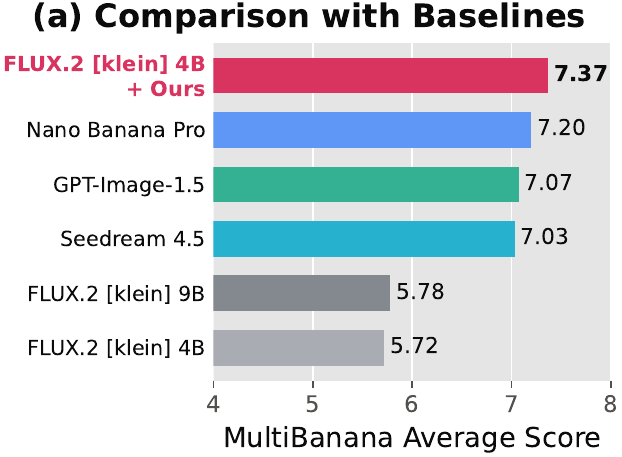}
    \end{minipage}
    \begin{minipage}{0.33\linewidth}
        \centering
        \includegraphics[width=\linewidth]{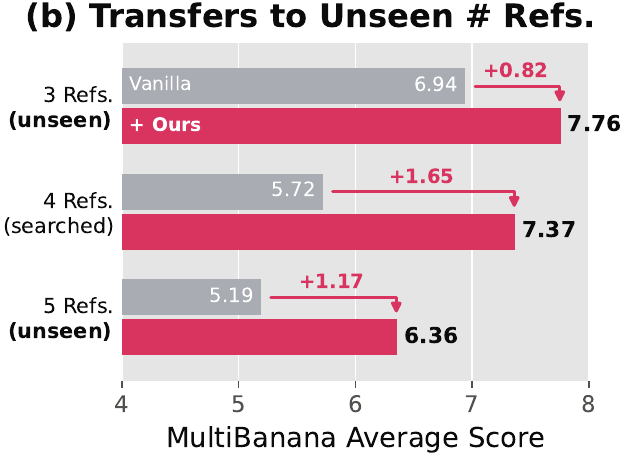}
    \end{minipage}\hfill
    \begin{minipage}{0.33\linewidth}
        \centering
        \includegraphics[width=\linewidth]{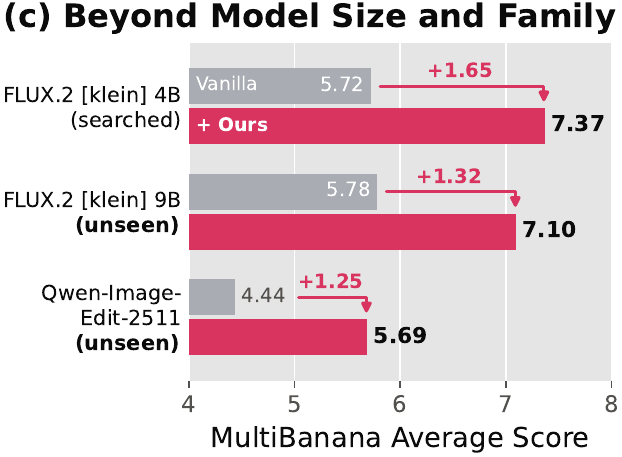}
    \end{minipage}\hfill
    \caption{
    AutoRef-Harness (+ Ours), optimized on four-reference MultiBanana tasks with FLUX.2 [klein] 4B, (\textbf{a}) enables FLUX.2 [klein] 4B to match or exceed proprietary models on the held-out test split and transfers unchanged to (\textbf{b}) unseen three- and five-reference settings and (\textbf{c}) generators of different scales and families.
    }
    \label{fig:teaser}
\end{figure}

\section{Introduction}
Recent image generation models can take multiple reference images as input and combine them into a new image~\citep{google2025nanobanana, google2025nanobananapro, openai2025gpt4oimage, wu2025qwenimagetechnicalreport}. 
This capability, referred to as multi-reference image generation~\citep{wu2025omnigen2, xia2025dreamomni2, zhang2026rcedit, huang2026scaling}, matters for practical image creation because users can specify people, objects, clothing, backgrounds, and styles using separate images. 
Such control is directly useful in applications including advertising~\citep{inoue2023layout, morita2025tkg}, virtual try-on~\citep{zhu2023tryondiffusion, chong2024catvton, cvpr2026garments2look}, and content creation~\citep{ruiz2022dreambooth, xu2026contextgen}. 
Yet combining multiple references correctly remains challenging. 
Models may omit or duplicate subjects from the references, or produce images in which the subjects appear pasted in rather than forming a coherent scene~\citep{xia2025dreamomni2, huang2026scaling}.

Recent work has proposed image generation agents that combine image generation and reasoning models through a harness and iteratively plan, generate, diagnose, and refine~\citep{hao2023promptist, yang2025idea2img, ma2025inferencetime, he2026gems}. 
The harness is executable code that specifies how the frozen models are used: how references are interpreted, prompts are constructed, candidates are generated and evaluated, and the output is selected. 
In multi-reference image generation, however, a good harness is harder to design than in text-to-image generation: references play different roles (e.g., identity, background, or style), and outputs must satisfy many criteria, including fidelity to each reference and the naturalness of the whole image~\citep{oshima2026multibanana, huang2026scaling}.
Many parts of the harness could therefore be improved, from which references to provide and in what order to how outputs are checked against each one, yet the effect of each change is hard to predict.
Indeed, existing human-written harnesses vary widely in performance (Section~\ref{sec:experiments_comparison_harness}).

Recent work on automatic agent optimization has expanded from prompts and workflows to executable code~\citep{lee2026metaharness,zhang2026dgm,miyai2026taskcoevolve}.
These methods, however, have largely been developed for tasks with verifiable rewards such as math~\citep{lee2026metaharness} and coding~\citep{lin2026agenticharness, zhang2026selfharnes}, whereas image generation relies on noisy visual evaluation and provides little diagnostic information through scalar scores alone. 
We therefore propose \textbf{AutoRef}, which optimizes harness code while keeping the image generation and reasoning models frozen. 
To address these challenges, AutoRef separates the tasks used to propose harness updates from those used to select candidates, so that selection does not reuse the examples the proposer sees.
It also runs an iterative beam search, in which the top-ranked harnesses on the selection tasks, rather than a harness chosen by the proposer, become the next parents.

\textbf{AutoRef-Harness}, the optimized harness for multi-reference image generation, was discovered by AutoRef on the MultiBanana benchmark~\citep{oshima2026multibanana} using the open-weight FLUX.2 [klein] 4B~\citep{blackforestlabs2026flux2klein}. 
It uses reference-grounded prompting, generates structurally diverse drafts, revises the better draft from explicit complaints, and selects among candidates with failure-aware comparisons.
With AutoRef-Harness, FLUX.2 [klein] 4B improves from 5.72 to 7.37 on the four-reference MultiBanana held-out test split, matching or exceeding proprietary models including Nano Banana Pro~\citep{google2025nanobananapro} and GPT-Image-1.5~\citep{openai2025gptimage1_5} (\autoref{fig:teaser}).
The same harness also improves results without re-optimization when the generator, number of references, benchmark, evaluator, or reasoning model is changed.
We release our code and AutoRef-Harness.

\section{Related Work}
\noindent\textbf{Multi-Reference Image Generation.}~~
Reference-conditioned image generation has evolved from personalized adaptation to specific subjects, as in DreamBooth~\citep{ruiz2022dreambooth}, toward general-purpose multimodal generation that incorporates multiple reference images. 
Recent models support flexible multi-reference image generation and editing~\citep{deng2025bagel, xia2025dreamomni2, wu2025omnigen2, blackforestlabs2026flux2klein, wu2025qwenimagetechnicalreport, 
google2025nanobanana, google2025nanobananapro, google2025nanobanana2, openai2025gpt4oimage, openai2025gptimage1_5}.
In parallel, recent work has improved multi-reference image generation by scaling reference-conditioned training data and fine-tuning the underlying models~\citep{zhang2026rcedit, huang2026scaling}.
In contrast, MultiBanana~\citep{oshima2026multibanana} shows that simple agentic refinement gives only limited gains on multi-reference tasks, indicating that simply wrapping a strong generator with a fixed agent workflow is insufficient. 
We therefore optimize the agent harness automatically from task feedback, improving multi-reference generation while keeping the generator frozen.

\noindent\textbf{Automatic Optimization of Agentic Systems.}~~
Automatic agent optimization searches over prompts~\citep{zhou2023ape,yang2024opro,pryzant2023protegi,guo2024evoprompt}, modular pipelines~\citep{khattab2024dspy,opsahlong2024mipro}, and agent workflows~\citep{zhuge2024gptswarm,hu2025adas,zhang2025aflow}.
Language-based feedback guides revisions \citep{yuksekgonul2025textgrad,agrawal2026gepa}, while program evolution extends optimization to executable code and self-improving agents \citep{novikov2025alphaevolve,lange2026shinkaevolve,zelikman2024stop,robeyns2025sica,zhang2026dgm}.
\citet{pryzant2023protegi} and \citet{guo2024evoprompt} keep multiple candidates across iterations, and \citet{agrawal2026gepa} and \citet{khattab2024dspy} can select them on held-out examples; all of them tune prompts within a fixed program.
Meta-Harness~\citep{lee2026metaharness} lets a coding agent read the code, scores, and execution trajectories of prior candidates, choose which one to build on, and revise the harness around a frozen model, scoring candidates on the same tasks that supply this feedback.
AutoDesign~\citep{luo2026autodesign} applies harness optimization to academic paper-to-poster generation.
\autoref{app:related_tts} discusses inference-time scaling for multimodal generation.

\section{Preliminaries}
\label{sec:prelim}
\noindent\textbf{Multi-Reference Image Generation.}~~
Let $x=(u,\mathcal{I})$ denote a multi-reference image generation task~\citep{wu2025omnigen2, xia2025dreamomni2, huang2026scaling}, where $u$ is the user prompt (the instruction) and $\mathcal{I}=\{I_1,\ldots,I_m\}$ is the set of reference images. Let $M_\theta$ and $G_\phi$ denote a frozen reasoning model and image generator, respectively. A harness $H$ is an executable program that specifies how these models are used: how references are interpreted, prompts constructed, candidates generated and evaluated, and the output selected. Running the harness yields an image $y$ and an execution trajectory $\tau$:
\begin{equation}
(y,\tau) \sim H(M_\theta,G_\phi,x).
\end{equation}

\noindent\textbf{Harness Optimization.}~~
Our goal is to optimize the harness while keeping the model parameters $\theta$ and $\phi$ frozen~\citep{zhang2026dgm, lin2026agenticharness, miyai2026taskcoevolve}. 
Let $p_{\mathrm{task}}$ denote the distribution of multi-reference image generation tasks, and let $R(y,x)$ denote an evaluator that scores the quality of a generated image $y$ for task $x$. We define the performance of a harness as
\begin{equation}
J(H)
=
\mathbb{E}_{x \sim p_{\mathrm{task}},\,
y \sim H(M_\theta, G_\phi, x)}
\left[
R(y,x)
\right].
\end{equation}

The harness optimization objective is therefore
\begin{equation}
H^\star = \arg\max_H J(H).
\label{eq:objective}
\end{equation}

With $M_\theta$ and $G_\phi$ frozen, optimization acts only on the executable code surrounding the models, which allows changes to prompting, generation, evaluation, selection, and control flow.

\noindent\textbf{Harness Search Loop.}~~
We approach this optimization problem through iterative code improvement~\citep{zhang2026dgm, lee2026metaharness, lin2026agenticharness}.
At iteration $t$, a coding-agent proposer $P$ has access to the current harness $H_t$ and an accumulated search history $\mathcal{L}_t$ of artifacts from previous iterations: harness implementations, evaluation scores, and execution trajectories.
The proposer can selectively inspect and search prior artifacts, diagnose failure modes, and decide how to modify the harness.
It then proposes an updated harness:
\begin{equation}
    H_{t+1} \leftarrow P(H_t, \mathcal{L}_t).
    \label{eq:search_loop}
\end{equation}
The proposed harness is evaluated on a set of search tasks, and its implementation, scores, and trajectories are added to the history $\mathcal{L}_{t+1}$.

\section{AutoRef}
\label{sec:method}
\begin{figure}[t]
\centering
\includegraphics[width=\textwidth]{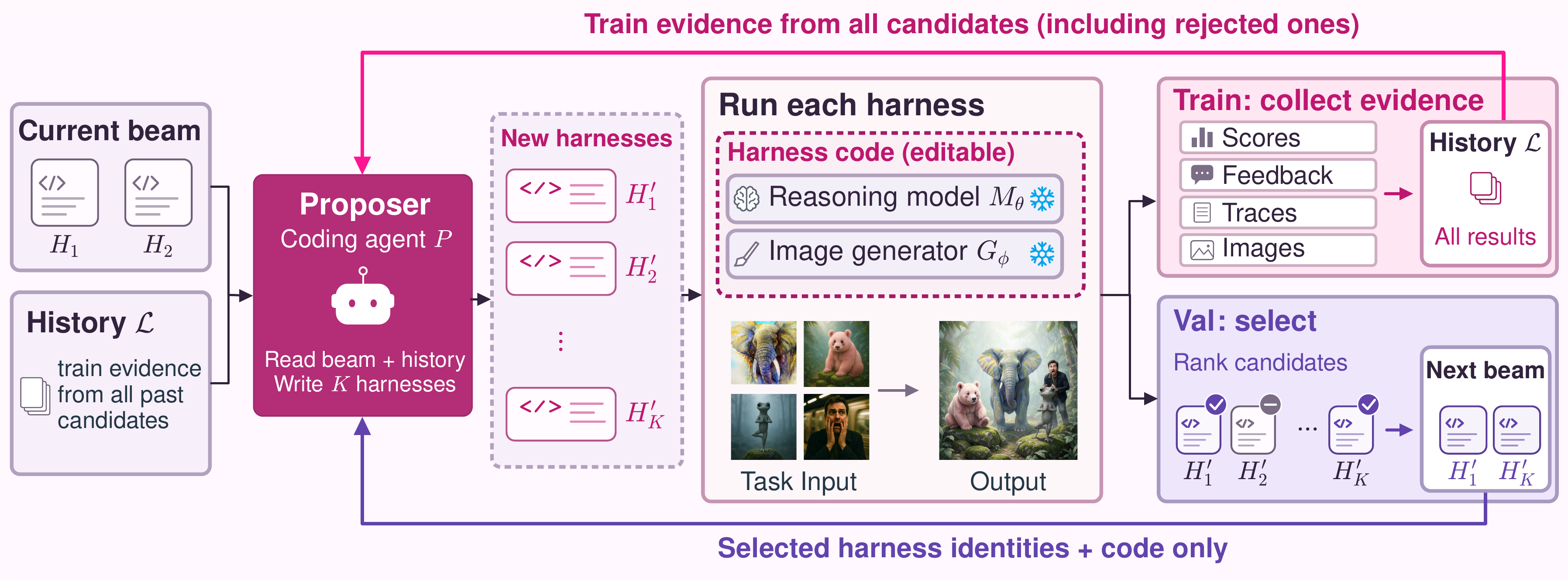}
\caption{\textbf{AutoRef}, one iteration. A harness is a program that calls a frozen reasoning model
$M_{\theta}$ and a frozen image generator $G_{\phi}$; the search rewrites this program. The proposer,
a coding agent $P$, reads the current beam and the search history $\mathcal{L}$ and writes $K=4$ new harnesses as code.
On $D_{\mathrm{train}}$ (pink), everything from
every candidate, rejected ones included --- scores, evaluator rationales, execution trajectories,
and generated images --- enters $\mathcal{L}$, where $P$ may inspect it. $D_{\mathrm{val}}$ (purple)
only ranks the four and keeps $B=2$; $P$ is told which two survived but never their scores. The
survivors form the next beam; their parents do not compete again.}
\label{fig:autoref}
\end{figure}

We propose \textbf{AutoRef}, a method for automatically optimizing harnesses for multi-reference image generation. 
Existing harness optimization methods~\citep{zhang2026dgm, lee2026metaharness, miyai2026taskcoevolve} primarily target tasks whose performance can be verified using discrete labels or executable tests. 
In image generation, however, a visual evaluator must estimate quality; failures are often hard to diagnose from scalar rewards alone, and repeated optimization over a limited set of evaluated examples can overfit to both the search tasks and the evaluator. 
AutoRef addresses these challenges by (1) separating the tasks used for harness updates from those used for candidate selection, and (2) using beam search that keeps the top-$B$ candidates on the validation tasks as parents for the next iteration. 
These choices adapt harness optimization to perceptual, non-verifiable image generation tasks.
\autoref{fig:autoref} illustrates one iteration; Algorithm~\ref{alg:meta_harness} (\autoref{app:algorithm}) gives the full procedure.

\noindent\textbf{Task Separation for Proposal and Selection.}~~
Directly optimizing against rich but non-verifiable evaluation feedback risks overfitting the harness to both a small set of search tasks and noise in the evaluator~\citep{huang2026envharness, luo2026autodesign}.
We therefore separate the tasks used to propose harness updates from those used to select among them.
We split the search tasks into disjoint sets $D_{\mathrm{train}}$ and $D_{\mathrm{val}}$, and write $J_{D}(H)$ for the mean of $R(y,x)$ over $x\in D$.
Evaluations on $D_{\mathrm{train}}$ provide feedback for harness improvement: scores, evaluator rationales, execution trajectories, and visual artifacts are added to the search history $\mathcal{L}$ and may be inspected by the proposer.
In contrast, $D_{\mathrm{val}}$ is used only for candidate selection, and its scores and artifacts are never exposed to the proposer or added to $\mathcal{L}$.
Thus, the proposer constructs new harnesses using only training-side feedback, while $J_{D_{\mathrm{val}}}$ selects among them without becoming a direct optimization signal.

\noindent\textbf{Iterative Beam Search.}~~
Selecting a single harness at each iteration can commit the search to a lineage favored by stochastic generation or noisy visual evaluation.
We therefore maintain a beam of $B$ harnesses.
At iteration $t$, the proposer uses the current beam and accumulated search history to generate $K$ candidate harnesses.
%
%
Each candidate is evaluated on both $D_{\mathrm{train}}$ and $D_{\mathrm{val}}$, and the next beam is formed by the $B$ candidates with the highest $J_{D_{\mathrm{val}}}$.
%
%
The proposer is told which candidates were selected but not their validation scores, while training-side evidence from all candidates, including unselected ones and their generated images, is preserved in $\mathcal{L}$ for subsequent iterations.
In our experiments, we use $B=2$ and $K=4$, and initialize the beam with two harnesses: the base generator $H_{\mathrm{vanilla}}$ (the generator called once on the user prompt) and GEMS~\citep{he2026gems} as $H_{\mathrm{init}}$. In the first iteration, the proposer writes all $K$ candidates from the two initial harnesses; in each later iteration, it writes two candidates from each beam member. The search that produced AutoRef-Harness is traced in \autoref{app:lineage}.

\section{The Optimized AutoRef-Harness}
\label{sec:optimized_harness}
\textbf{AutoRef-Harness} is the harness returned by AutoRef (Section~\ref{sec:method}). It draws three
images from $G_\phi$ and makes all other decisions with $M_\theta$: it generates drafts A and B from two
differently structured prompts, keeps the better one, generates draft C from complaints about the
winner, and returns the better of the winner and C (\autoref{app:harness-flow}).
Compared with human-written harnesses, it differs in how each step is specialized for multiple
references and how the steps are chained: every prompt assigns each requested element to its
reference (\S\ref{sec:optimized_grounding}); the two drafts differ in prompt structure, not only in
sampling (\S\ref{sec:optimized_generation}); complaints name the reference they concern
(\S\ref{sec:optimized_complaint}); and selection counts hard failures (e.g., a missing reference)
before pairwise judgment, and a later draft replaces the incumbent only if it wins under this
rule (\S\ref{sec:optimized_selection}). Each component was added in an iteration that
raised the validation score, and alternatives like editing the winner in place were dropped
(\autoref{app:lineage}).

\subsection{Reference-Grounded Prompting}
\label{sec:optimized_grounding}
References play different roles (identity, garment, attribute, background, style); a generator
that confuses them leaks attributes or drops references. AutoRef-Harness never
passes the raw instruction to the generator: $M_\theta$ reads the instruction and all references
and writes a prompt that assigns each requested element to its reference, excluding unrequested
content; later prompts use the same format.

\subsection{Structurally Diverse Drafts}
\label{sec:optimized_generation}
A common failure is a pasted-in look: each subject matches its reference, but its lighting,
perspective, or colors disagree with the scene. Resampling one prompt rarely fixes this, so the two
drafts use different prompt structures. Draft A describes the scene subject by subject. For draft B,
$M_\theta$ identifies the reference designated as the background or style, and the prompt asks the
generator to keep that reference as the canvas and paint the other subjects into it, so that
subjects and scene are rendered jointly. If no such reference exists, draft B is a second sample of
draft A's prompt.

\subsection{Complaint-Directed Revision}
\label{sec:optimized_complaint}
$M_\theta$ lists up to five concrete complaints about the winner of A and B, each naming the
reference it concerns (e.g., wrong identity, attribute from the wrong reference, inconsistent
lighting), and rewrites the prompt to address them; draft C is generated from the revised prompt,
or by resampling the winner's prompt if there is no complaint.

\subsection{Failure-Aware Selection}
\label{sec:optimized_selection}
Candidates are compared in pairs. Each draft is first checked for hard failures (missing
reference, extra or duplicated subject, wrong background); the draft with fewer failures wins. On a tie, $M_\theta$ lists the differences a strict rater would score and names a winner in
both presentation orders; the challenger (draft B, then draft C) must win both. Selection uses $M_\theta$ (GPT-5.5), not the evaluator $R$.

\section{Experiments}
\subsection{Experimental Settings}

\noindent\textbf{Benchmarks.}~~We evaluate on MultiBanana~\citep{oshima2026multibanana}, a benchmark for multi-reference image generation.
We use the 229 tasks with four reference images, split into 48 training tasks, 48 validation tasks, and 133 test tasks, with Qwen3-VL-8B-Instruct~\citep{bai2025qwen3vl} as the evaluator.
We use the training and validation splits for harness optimization, while the held-out test split remains unseen during search.
To test generalization across unseen reference counts, we further evaluate on the three- and five-reference settings, randomly sampling 24 tasks per task type (96 per setting).

To evaluate generalization beyond the benchmark and evaluator, we also test on OmniContext~\citep{wu2025omnigen2}, which we never use during harness search.
We randomly sample 15 tasks from each task type, for 120 tasks in total, and evaluate them using the official GPT-4.1~\citep{openai2023gpt4} evaluator.
With both the benchmark and the evaluator differing from those used in the search, this setting tests whether the learned harness transfers to unseen data distributions and evaluation signals.

\noindent\textbf{Harness Search.}~~We initialize the search with two harnesses: the base FLUX.2 [klein] 4B generator~\citep{blackforestlabs2026flux2klein} and GEMS~\citep{he2026gems}, an image-generation harness configured with FLUX.2 [klein] 4B as the generator and GPT-5.5~\citep{openai2026gpt55} as the reasoning model.
For AutoRef (Section~\ref{sec:method}), we use Claude Fable 5.1 as the proposer through the Claude Code CLI~\citep{anthropic2025claudecode} and run five search iterations.
Implementation details and model versions are in \autoref{app:impl}, and the proposer's prompts are in \autoref{app:prompts}.

\noindent\textbf{Baselines.}~~We compare against a broad set of baselines: proprietary image models including GPT-Image-1.5~\citep{openai2025gptimage1_5}, Nano Banana Pro~\citep{google2025nanobananapro}, and Seedream 4.5~\citep{bytedance2025seedream45}, 
open image models including OmniGen2~\citep{wu2025omnigen2}, DreamOmni2~\citep{xia2025dreamomni2}, BAGEL~\citep{deng2025bagel}, FLUX.2 [klein] 4B and 9B~\citep{blackforestlabs2026flux2klein}, and Qwen-Image-Edit-2511~\citep{wu2025qwenimagetechnicalreport}, 
and agentic or search-based methods including Best-of-$N$~\citep{ma2025inferencetime}, GEMS~\citep{he2026gems}, IPR~\citep{oshima2026multibanana}, and Idea2Img~\citep{yang2025idea2img}.
We also report the harness that Meta-Harness~\citep{lee2026metaharness} converges to under the same budget, generator, reasoning model, and evaluator, so the search algorithm is the only difference between it and AutoRef-Harness.

\begin{table}[t]
\centering
\caption{MultiBanana~\citep{oshima2026multibanana} with 4 references, held-out test split. 
Cells are the mean of the five evaluation metrics (1--10) per task type and Avg.\ the mean of the four.
\emph{Gen.}\ is images drawn per task, a measured mean over the split rather than a nominal maximum. Best open image generators per column in bold, second best underlined.
$^\dagger$: AutoRef-Harness discovered with FLUX.2 [klein] 4B on the four-reference MultiBanana setting.}
\label{tab:multibanana-summary-4ref}
\resizebox{0.8\columnwidth}{!}{%
\begin{tabular}{l c ccccc}
\toprule
Method & Gen. & Object & Local & Global & Background & \textbf{Avg.} \\
\midrule
\multicolumn{7}{c}{Proprietary Models} \\
\midrule
GPT-Image-1.5 & 1 & 6.84 & 8.12 & 7.22 & 6.11 & 7.07 \\
Nano Banana Pro & 1 & 6.73 & 7.98 & 7.70 & 6.39 & 7.20 \\
Seedream 4.5 & 1 & 6.65 & 7.54 & 7.83 & 6.10 & 7.03 \\
\midrule
\multicolumn{7}{c}{Open Models} \\
\midrule
OmniGen2 & 1 & 3.56 & 4.35 & 3.64 & 3.43 & 3.75 \\
DreamOmni2 & 1 & 3.38 & 5.17 & 3.51 & 3.28 & 3.83 \\
BAGEL & 1 & 2.64 & 4.16 & 2.76 & 3.04 & 3.15 \\
\midrule
FLUX.2 [klein] 4B & 1 & 5.95 & 6.11 & 5.50 & 5.34 & 5.72 \\
\pinkcell{\quad $+$ \textbf{AutoRef-Harness}}
    & \pinkcell{3}
    & \pinkcell{\underline{7.27}}
    & \pinkcell{\textbf{7.87}}
    & \pinkcell{\textbf{7.64}}
    & \pinkcell{\textbf{6.70}}
    & \pinkcell{\textbf{7.37}} \\
FLUX.2 [klein] 9B & 1 & 6.80 & 5.64 & 5.58 & 5.13 & 5.78 \\
\pinkcell{\quad $+$ \textbf{AutoRef-Harness}$^\dagger$}
    & \pinkcell{3}
    & \pinkcell{\textbf{7.49}}
    & \pinkcell{\underline{7.30}}
    & \pinkcell{\underline{7.22}}
    & \pinkcell{\underline{6.37}}
    & \pinkcell{\underline{7.10}} \\
Qwen-Image-Edit-2511 & 1 & 4.05 & 5.05 & 4.43 & 4.23 & 4.44 \\
\pinkcell{\quad $+$ \textbf{AutoRef-Harness}$^\dagger$}
    & \pinkcell{3}
    & \pinkcell{5.50}
    & \pinkcell{6.07}
    & \pinkcell{5.79}
    & \pinkcell{5.41}
    & \pinkcell{5.69} \\
\bottomrule
\end{tabular}
}
\end{table}

\subsection{Main Results}

\label{sec:main_results}

\begin{table*}[t]
\centering
\caption{MultiBanana with 3 and 5 references. 
AutoRef-Harness largely preserves its performance gains when transferred to unseen reference counts.
Best open image generators per column in bold, second best underlined.
$^\dagger$: AutoRef-Harness discovered with FLUX.2 [klein] 4B on the four-reference MultiBanana setting.
}
\label{tab:multibanana-summary-3-5ref}
\resizebox{\textwidth}{!}{%
\begin{tabular}{l c ccccc ccccc}
\toprule
 & & \multicolumn{5}{c}{3 references} & \multicolumn{5}{c}{5 references} \\
\cmidrule(lr){3-7} \cmidrule(lr){8-12}
Method & Gen. & Object & Local & Global & Backg. & \textbf{Avg.} & Object & Local & Global & Backg. & \textbf{Avg.} \\
\midrule
\multicolumn{12}{c}{Proprietary Models} \\
\midrule
GPT-Image-1.5 & 1 & 8.42 & 8.14 & 7.84 & 7.03 & 7.86 & 5.68 & 8.48 & 6.20 & 6.02 & 6.60 \\
Nano Banana Pro & 1 & 8.27 & 8.09 & 7.62 & 6.03 & 7.50 & 5.53 & 8.41 & 7.03 & 5.88 & 6.71 \\
Seedream 4.5 & 1 & 6.88 & 8.43 & 7.68 & 6.67 & 7.41 & 6.02 & 8.16 & 6.19 & 6.04 & 6.60 \\
\midrule
\multicolumn{12}{c}{Open Models} \\
\midrule
OmniGen2 & 1 & 4.74 & 5.78 & 5.62 & 4.87 & 5.25 & 2.68 & 4.37 & 2.73 & 2.93 & 3.18 \\
DreamOmni2 & 1 & 5.31 & 6.26 & 5.20 & 4.68 & 5.36 & 2.01 & 4.88 & 3.17 & 2.42 & 3.12 \\
BAGEL & 1 & 4.47 & 4.50 & 3.69 & 3.72 & 4.10 & 2.20 & 4.16 & 2.83 & 2.08 & 2.82 \\
\midrule
FLUX.2 [klein] 4B & 1 & 6.26 & 7.90 & 6.83 & 6.78 & 6.94 & 4.04 & 6.32 & 5.72 & 4.70 & 5.19 \\
\pinkcell{\quad $+$ \textbf{AutoRef-Harness}$^\dagger$} & \pinkcell{3} & \pinkcell{\underline{7.15}} & \pinkcell{\textbf{8.68}} & \pinkcell{\underline{7.82}} & \pinkcell{\textbf{7.41}} & \pinkcell{\underline{7.76}} & \pinkcell{\underline{5.67}} & \pinkcell{7.37} & \pinkcell{\textbf{6.75}} & \pinkcell{\underline{5.65}} & \pinkcell{\underline{6.36}} \\
FLUX.2 [klein] 9B & 1 & 6.89 & 7.36 & 6.69 & 6.36 & 6.83 & 4.84 & \underline{7.49} & 5.64 & 4.45 & 5.61 \\
\pinkcell{\quad $+$ \textbf{AutoRef-Harness}$^\dagger$}
    & \pinkcell{3}
    & \pinkcell{\textbf{8.04}}
    & \pinkcell{\underline{8.29}}
    & \pinkcell{\textbf{7.87}}
    & \pinkcell{\underline{7.09}}
    & \pinkcell{\textbf{7.82}}
    & \pinkcell{\textbf{6.16}}
    & \pinkcell{\textbf{7.67}}
    & \pinkcell{\underline{6.12}}
    & \pinkcell{\textbf{5.92}}
    & \pinkcell{\textbf{6.47}} \\
Qwen-Image-Edit-2511 & 1 & 4.56 & 5.13 & 5.53 & 5.03 & 5.06 & 1.35 & 1.83 & 2.56 & 1.80 & 1.89 \\
\pinkcell{\quad $+$ \textbf{AutoRef-Harness}$^\dagger$}
    & \pinkcell{3}
    & \pinkcell{6.87}
    & \pinkcell{7.86}
    & \pinkcell{6.67}
    & \pinkcell{6.39}
    & \pinkcell{6.95}
    & \pinkcell{1.57}
    & \pinkcell{1.69}
    & \pinkcell{2.30}
    & \pinkcell{1.93}
    & \pinkcell{1.87} \\
\bottomrule
\end{tabular}}
\end{table*}

\begin{figure}[t]
\centering
\includegraphics[width=\textwidth]{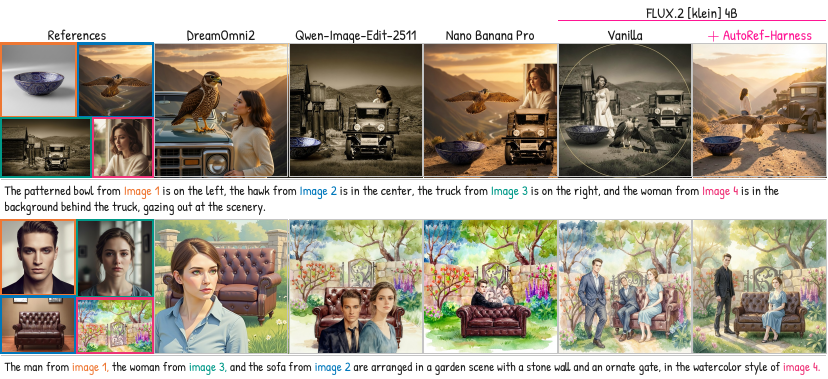}
\caption{Qualitative comparison on held-out four-reference MultiBanana tasks.
Each reference is outlined in a distinct color, and the phrase in the instruction that refers to it is shown in the same color.
Baseline models often omit, duplicate, or misplace references, or copy and paste them unnaturally, whereas AutoRef-Harness preserves each reference and integrates all four into a coherent scene.}
\label{fig:qualitative}
\end{figure}

As shown in \autoref{tab:multibanana-summary-4ref}, AutoRef-Harness improves the performance of FLUX.2 [klein] 4B on the four-reference MultiBanana held-out test split.
Despite using the relatively small FLUX.2 [klein] 4B as its image generator, the resulting system outperforms all evaluated open models and achieves performance competitive with proprietary models such as GPT-Image-1.5, Nano Banana Pro, and Seedream 4.5.
The gains from AutoRef-Harness also transfer beyond the model used during harness optimization: applying the same harness to the larger FLUX.2 [klein] 9B improves its performance, and replacing FLUX.2 with Qwen-Image-Edit-2511 likewise yields a substantial gain.
These results indicate that the benefit of the discovered harness is not limited to a particular model scale or generator family.
Importantly, we achieve these improvements without updating the image generator or reasoning model; we only change the inference-time harness.
Per-metric results are reported in Appendix~\ref{sec:multibanana_per_metric}, and the harness further improves Qwen-Image-Edit-2511 after fine-tuning for multi-reference image generation with DyRef (\citealp{huang2026scaling}; Appendix~\ref{app:finetune}).

\autoref{fig:qualitative} illustrates qualitative examples.
Baseline models often omit, duplicate, or misplace references, or paste them in unnaturally.
For instance, the base FLUX.2 [klein] 4B duplicates the hawk and places the woman in the foreground rather than the background in the first example, and duplicates the man in the second.
Nano Banana Pro and Qwen-Image-Edit-2511 instead produce copy-and-paste-like results in the first and second examples, respectively.
AutoRef-Harness preserves each reference and naturally integrates it into the requested scene.
\autoref{app:qualitative} shows more examples.

\subsection{Transferability of AutoRef-Harness}

\noindent\textbf{Across Reference Counts.}~~AutoRef-Harness is discovered on four-reference MultiBanana tasks but applies to different reference counts without modification.
As shown in \autoref{tab:multibanana-summary-3-5ref}, it consistently improves FLUX.2 [klein] 4B on both the unseen three- and five-reference settings.
With three references, the resulting system outperforms all open image generators without the harness and remains competitive with proprietary models; the gain also persists in the more challenging five-reference setting.
The harness likewise improves the other generators, except Qwen-Image-Edit-2511 with five references, where the generator itself fails and the harness cannot compensate (\autoref{app:breakdown}).
AutoRef-Harness thus does not rely on the four-reference structure used during harness discovery.

\noindent\textbf{Across Benchmarks and Evaluators.}~~We further evaluate the same AutoRef-Harness on OmniContext, which is never used during harness discovery and is scored by its official GPT-4.1~\citep{openai2023gpt4} evaluator rather than the Qwen3-VL-8B-Instruct evaluator used for MultiBanana.
As shown in \autoref{tab:omnicontext-overall}, AutoRef-Harness again improves FLUX.2 [klein] 4B and remains competitive with proprietary image models.
Thus, the gains persist under simultaneous changes in both the benchmark distribution and the evaluator, without benchmark-specific modifications to the harness.

\begin{figure}[t]
    \centering
    \begin{minipage}{0.275\linewidth}
        \centering
        \includegraphics[width=\linewidth]{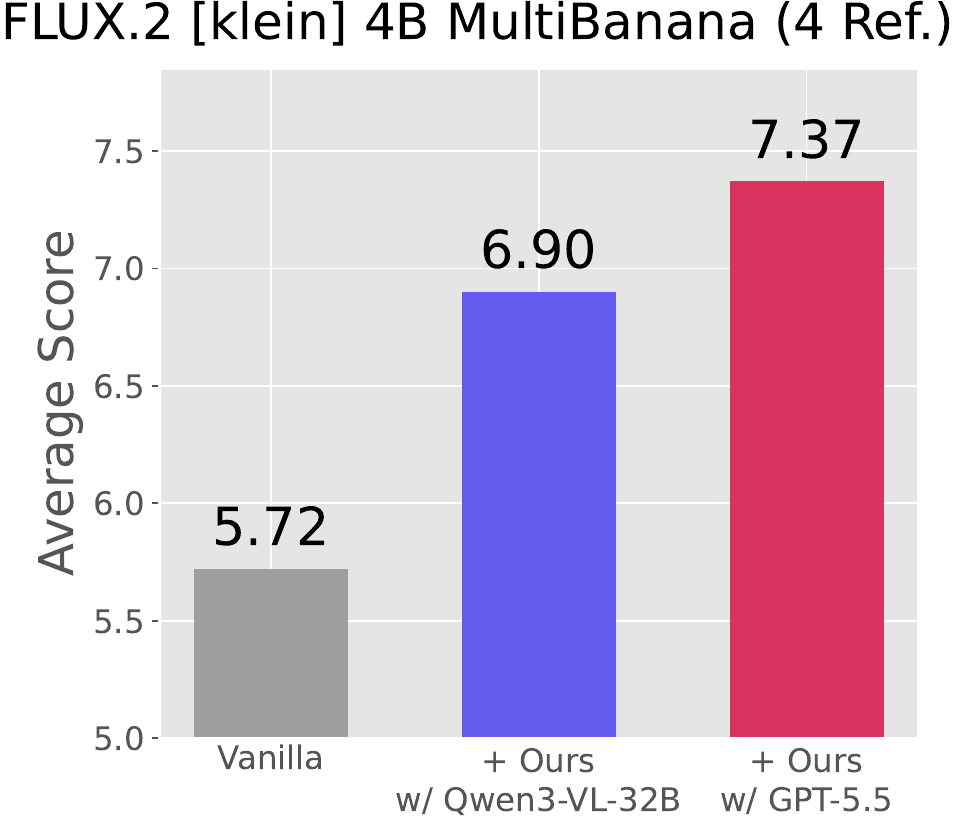}
    \end{minipage}\hfill
    \begin{minipage}{0.36\linewidth}
        \centering
        \includegraphics[width=\linewidth]{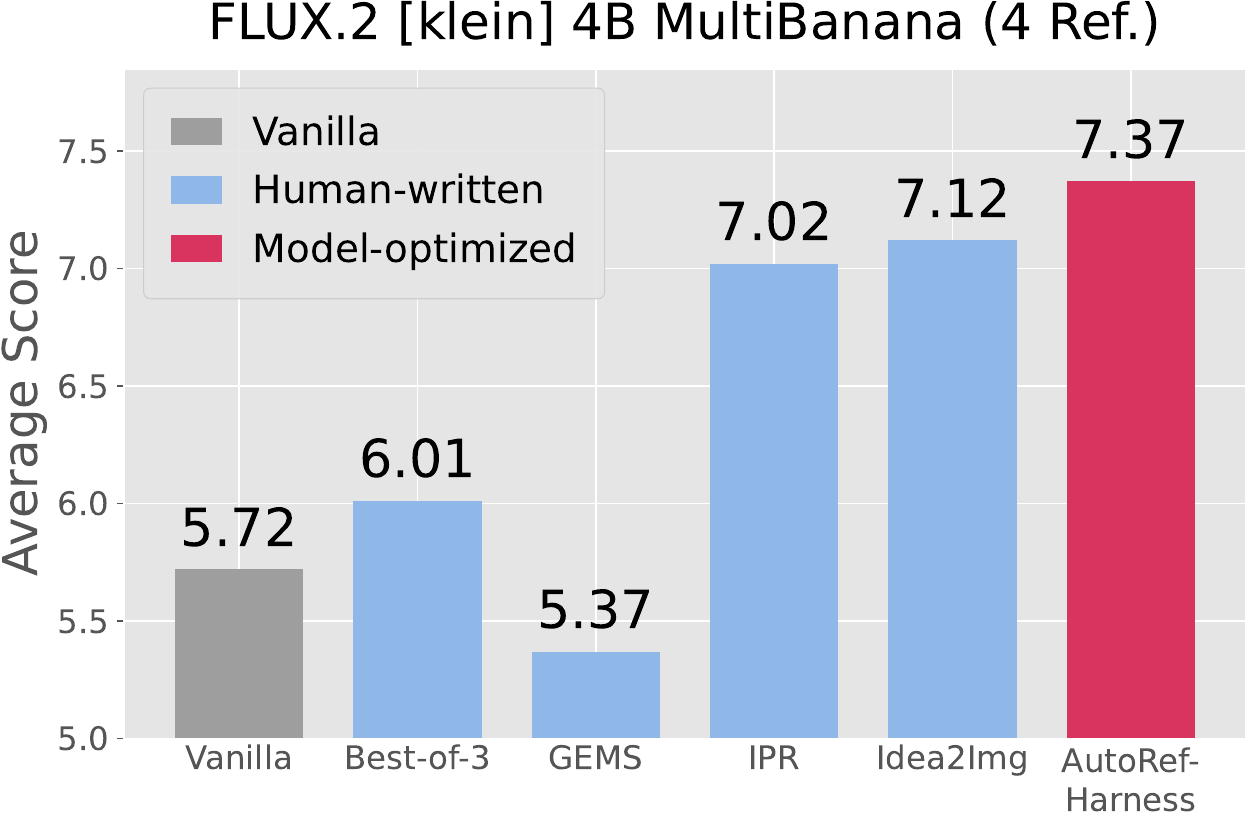}
    \end{minipage}
    \begin{minipage}{0.35\linewidth}
        \centering
        \includegraphics[width=\linewidth]{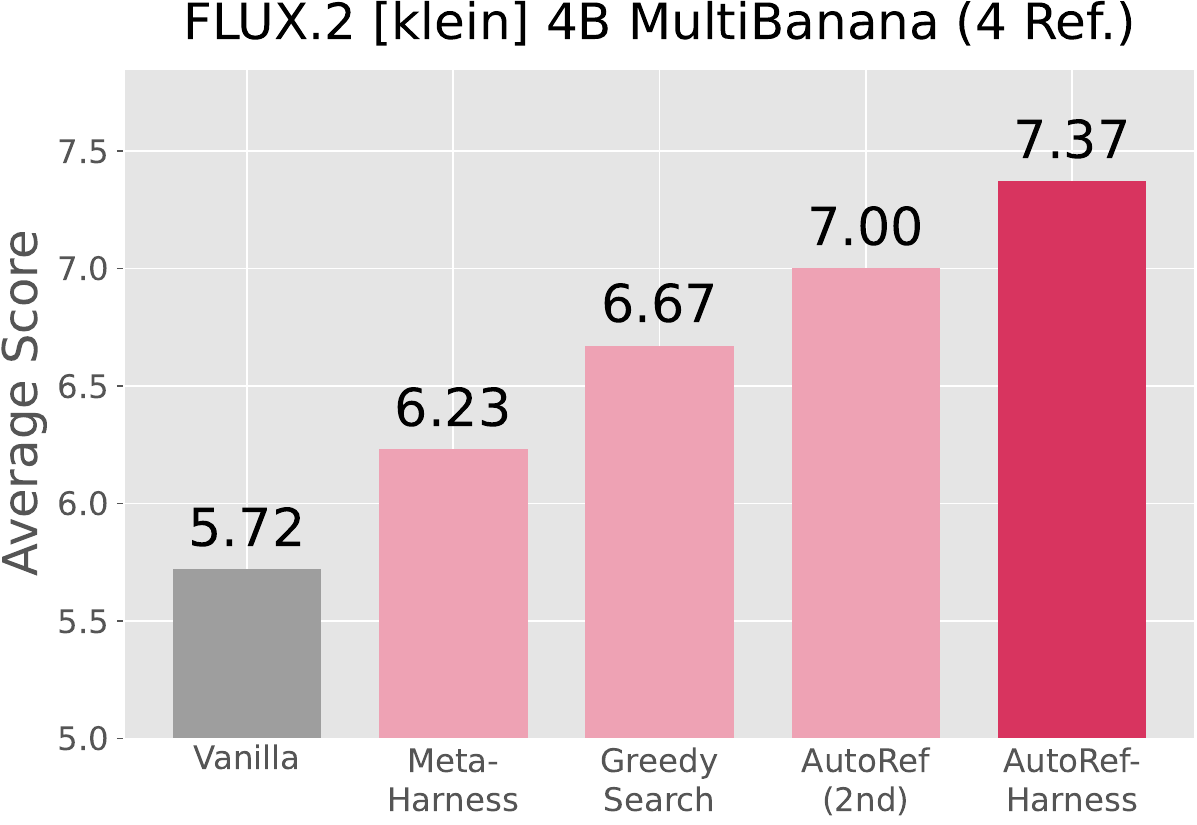}
    \end{minipage}\hfill
    \caption{
    (\textbf{Left}) Reasoning-model transfer.
    AutoRef-Harness (+ Ours) improves FLUX.2 [klein] 4B with both GPT-5.5 and the open-weight Qwen3-VL-32B.
    (\textbf{Middle}) Comparison with human-written harnesses.
    AutoRef-Harness outperforms Best-of-3, GEMS, IPR, and Idea2Img.
    (\textbf{Right}) Comparison of harness optimization methods.
    AutoRef discovers stronger harnesses than Meta-Harness and Greedy Search; even its second-best discovered harness outperforms the final harnesses produced by both alternative search procedures.
    }
    \label{fig:bar_comparison}
\end{figure}

\noindent\textbf{Across Reasoning Models.}~~As shown in Section~\ref{sec:main_results}, AutoRef-Harness transfers across image generators of different scales and families.
We next test whether the same harness also transfers across reasoning models.
With the open-weight Qwen3-VL-32B~\citep{bai2025qwen3vl} in place of GPT-5.5 and the harness structure unchanged, AutoRef-Harness improves FLUX.2 [klein] 4B from 5.72 to 6.90 on the four-reference MultiBanana held-out test split (7.37 with GPT-5.5; \autoref{fig:bar_comparison}, \textbf{Left}).
This suggests that the orchestration strategy encoded by AutoRef-Harness is not specific to the proprietary reasoning model used during its discovery.

\begin{table*}[t]
\centering
\caption{OmniContext~\citep{wu2025omnigen2}, unseen during the search and scored by its GPT-4.1 evaluator.
Cells are the geometric mean of prompt following and subject consistency (0--10) per task type, and Avg.\ the mean of the eight; Char.: character, Obj.: object, C.$+$O.: character and object.
Best open image generators per column in bold, second best underlined.
$^\dagger$: AutoRef-Harness discovered with FLUX.2 [klein] 4B on the four-reference MultiBanana setting.
}
\label{tab:omnicontext-overall}
\resizebox{\textwidth}{!}{%
\begin{tabular}{l c cc ccc ccc c}
\toprule
 & & \multicolumn{2}{c}{SINGLE} & \multicolumn{3}{c}{MULTIPLE} & \multicolumn{3}{c}{SCENE} & \\
\cmidrule(lr){3-4} \cmidrule(lr){5-7} \cmidrule(lr){8-10}
Method & Gen. & Char. & Obj. & Char. & Obj. & C.$+$O. & Char. & Obj. & C.$+$O. & \textbf{Avg.} \\
\midrule
\multicolumn{11}{c}{Proprietary Models} \\
\midrule
GPT-Image-1.5 & 1 & 9.56 & 9.70 & 9.32 & 9.46 & 9.26 & 9.69 & 9.39 & 8.93 & 9.41 \\
Nano Banana Pro & 1 & 9.63 & 9.42 & 9.46 & 9.19 & 9.02 & 9.35 & 8.39 & 8.20 & 9.08 \\
Seedream 4.5 & 1 & 9.45 & 9.50 & 9.09 & 9.39 & 9.09 & 9.35 & 8.66 & 8.23 & 9.09 \\
\midrule
\multicolumn{11}{c}{Open Models} \\
\midrule
OmniGen2 & 1 & 8.51 & 5.73 & 6.30 & 6.58 & 7.71 & 6.93 & 6.10 & 7.03 & 6.86 \\
DreamOmni2 & 1 & 7.81 & 6.72 & 4.80 & 7.07 & 5.92 & 5.78 & 5.63 & 5.72 & 6.18 \\
BAGEL & 1 & 6.67 & 7.09 & 3.43 & 6.74 & 7.08 & 3.97 & 4.11 & 5.63 & 5.59 \\
\midrule
FLUX.2 [klein] 4B & 1 & \textbf{9.22} & 8.16 & 7.91 & 8.21 & 8.68 & 9.18 & 7.47 & 7.54 & 8.30 \\
\pinkcell{\quad $+$ \textbf{AutoRef-Harness}$^\dagger$}
    & \pinkcell{3}
    & \pinkcell{8.94}
    & \pinkcell{8.99}
    & \pinkcell{9.02}
    & \pinkcell{8.78}
    & \pinkcell{8.59}
    & \pinkcell{\underline{9.28}}
    & \pinkcell{\textbf{8.89}}
    & \pinkcell{\underline{8.28}}
    & \pinkcell{\underline{8.85}} \\
FLUX.2 [klein] 9B & 1 & \underline{9.17} & 9.08 & 8.66 & 8.14 & \underline{8.74} & \textbf{9.35} & 8.07 & 7.47 & 8.59 \\
\pinkcell{\quad $+$ \textbf{AutoRef-Harness}$^\dagger$}
    & \pinkcell{3}
    & \pinkcell{\textbf{9.22}}
    & \pinkcell{\underline{9.18}}
    & \pinkcell{\textbf{9.19}}
    & \pinkcell{\textbf{9.32}}
    & \pinkcell{\textbf{8.82}}
    & \pinkcell{\textbf{9.35}}
    & \pinkcell{8.40}
    & \pinkcell{\textbf{8.30}}
    & \pinkcell{\textbf{8.97}} \\
Qwen-Image-Edit-2511 & 1 & 9.14 & \textbf{9.19} & 8.66 & \underline{9.00} & 8.33 & 6.97 & 8.17 & 8.17 & 8.45 \\
\pinkcell{\quad $+$ \textbf{AutoRef-Harness}$^\dagger$}
    & \pinkcell{3}
    & \pinkcell{9.12}
    & \pinkcell{8.64}
    & \pinkcell{\underline{9.09}}
    & \pinkcell{8.62}
    & \pinkcell{8.34}
    & \pinkcell{8.71}
    & \pinkcell{\underline{8.82}}
    & \pinkcell{\underline{8.28}}
    & \pinkcell{8.70} \\
\bottomrule
\end{tabular}}
\end{table*}

\subsection{Comparison with Harnesses and Harness Search}
\label{sec:experiments_comparison_harness}

\noindent\textbf{Comparison with Human-Written Harnesses.}~~We compare AutoRef-Harness with human-written harnesses: Best-of-3~\citep{ma2025inferencetime}, GEMS~\citep{he2026gems}, IPR~\citep{oshima2026multibanana}, and Idea2Img~\citep{yang2025idea2img}, which use 3, 2.7, 3, and 9 image generations per task, respectively (details in Appendix~\ref{app:baselines_human_written_harness}).
AutoRef-Harness achieves the highest performance on the four-reference MultiBanana held-out test split (\autoref{fig:bar_comparison}, \textbf{Middle}).
Notably, Idea2Img still underperforms despite using three times AutoRef-Harness's generation budget.

\noindent\textbf{Comparison with Harness Optimization Methods.}~~We next compare AutoRef with alternative harness optimization methods (details in Appendix~\ref{app:baselines_harness_search}).
Meta-Harness~\citep{lee2026metaharness} uses the same tasks for optimization feedback and candidate ranking, and its proposer chooses which candidate to build on from the full search history.
Greedy Search adopts AutoRef's train--validation separation but retains only the single best harness per iteration, whereas AutoRef keeps the top-$B$ candidates on $D_{\mathrm{val}}$ as parents for the next iteration.
Under the same generator, reasoning model, and evaluator, AutoRef yields the strongest final harness on the held-out test split, followed by Greedy Search and Meta-Harness (\autoref{fig:bar_comparison}, \textbf{Right}).
The final harnesses of Meta-Harness and Greedy Search also draw more images per task than AutoRef-Harness (4.2 and 5 vs.\ 3).
Moreover, even AutoRef's second-best harness outperforms both methods' final harnesses.
The successive gains from Meta-Harness to Greedy Search to AutoRef support the value of both train--validation separation and a multi-parent beam when optimizing image-generation harnesses from perceptual evaluation.

\subsection{Ablation Study}

\begin{wraptable}{r}{0.44\textwidth}
\vspace{-12pt}
\centering\small\setlength{\tabcolsep}{3.5pt}
\caption{Component ablation on the held-out test split. Columns are the components of
\S\ref{sec:optimized_grounding}--\ref{sec:optimized_selection}; parentheses give the change from the
full harness.}
\label{tab:ablation}
\resizebox{\linewidth}{!}{
\begin{tabular}{l cccc l}
\toprule
 & \S\ref{sec:optimized_grounding} & \S\ref{sec:optimized_generation} & \S\ref{sec:optimized_complaint} & \S\ref{sec:optimized_selection} & Avg. \\
\midrule
\pinkcell{\textbf{Full harness}} & \pinkcell{\yes} & \pinkcell{\yes} & \pinkcell{\yes} & \pinkcell{\yes} & \pinkcell{\textbf{7.37}} \\
\midrule
\multirow{3}{*}{Leave one out} & \yes & \no & \yes & \yes & 7.01 {\scriptsize($-$0.36)} \\
 & \yes & \yes & \no & \yes & 6.99 {\scriptsize($-$0.38)} \\
 & \yes & \yes & \yes & \no & 6.93 {\scriptsize($-$0.44)} \\
\midrule
Grounding only & \yes & \no & \no & \no & 6.91 {\scriptsize($-$0.46)} \\
Selection only & \no & \no & \no & \yes & 6.15 {\scriptsize($-$1.22)} \\
Generator only & \no & \no & \no & \no & 5.72 {\scriptsize($-$1.65)} \\
\bottomrule
\end{tabular}
}
\vspace{-8pt}
\end{wraptable}
We ablate one component at a time, replacing a removed draft or revision with another sample from the same prompt and failure-aware selection with the Best-of-$N$ selector (Appendix~\ref{app:baselines_human_written_harness}).
Reference-grounded prompting cannot be ablated this way, since all drafts and the revision use grounded prompts; we instead evaluate it alone with a single image (grounding only) and the harness without it (selection only).
\autoref{tab:ablation} shows that each of the four components contributes: removing structurally diverse drafts, complaint-directed revision, or failure-aware selection lowers the average score, and both grounding only and selection only improve over the generator alone.
Grounding only (6.91, one image) approaches IPR (7.02, three images), which also rewrites the prompt from the references, and selection only (6.15) is close to Best-of-3 (6.01), which differs only in its selector; yet the same selector adds 0.44 within the full harness (7.37 vs.\ 6.93), and no partial combination matches the full harness.

\subsection{Human Evaluation}

\begin{wrapfigure}{r}{0.42\linewidth}
\vspace{-12pt}
\centering
\includegraphics[width=\linewidth]{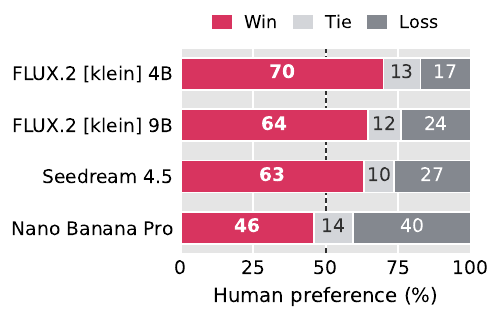}
\vspace{-16pt}
\caption{Win, tie, and loss rates (\%) of AutoRef-Harness (FLUX.2 [klein] 4B) against each baseline in pairwise human evaluation.}
\label{fig:human_eval}
\vspace{-10pt}
\end{wrapfigure}
AutoRef optimizes against an automatic evaluator, so we test whether its gains hold for human raters.
Four raters compared FLUX.2 [klein] 4B + AutoRef-Harness with each of four baselines on 50 tasks each from the four-reference held-out test split.
For each task, raters see the references, the instruction, and the two outputs in random order, and select the better output, or a tie only if they cannot distinguish the two.
As shown in \autoref{fig:human_eval}, AutoRef-Harness achieves a 70\% win rate against its base generator, FLUX.2 [klein] 4B (17\% loss), wins more often than it loses against FLUX.2 [klein] 9B (64\% vs.\ 24\%) and Seedream 4.5 (63\% vs.\ 27\%), and is competitive with Nano Banana Pro (46\% vs.\ 40\%).

\section{Discussion and Limitations}
AutoRef-Harness changes only how a frozen generator is used, so it cannot exceed what the generator can produce: when no draft is acceptable, as for Qwen-Image-Edit-2511 at five references, better selection does not help (\autoref{app:breakdown}).
The search maximizes a single VLM evaluator's score, yet the harness transfers to OmniContext and its GPT-4.1 evaluator.
As the evaluator and the proposer are replaceable, AutoRef may benefit from stronger VLMs and coding agents, and richer evaluators could be explored, e.g., combining a VLM with segmentation models~\citep{carion2025sam3segmentconcepts}.

\section{Conclusion}
We introduced AutoRef, a harness optimization method for multi-reference image generation that keeps the image generator and the reasoning model frozen and changes only the program that combines them.
AutoRef separates the tasks that inform proposals from those used to select among them, and continues from the top-ranked harnesses on the selection tasks through iterative beam search.
The harness it discovers, AutoRef-Harness, makes the open-weight FLUX.2 [klein] 4B competitive with proprietary models on MultiBanana, transfers to unseen reference counts, an unseen benchmark and evaluator, other generators, and another reasoning model, and adds to the gains of fine-tuning.
These results suggest that how frozen models are used is itself worth optimizing.

\subsection*{AI use statement}


In this work, we used generative AI tools for the following tasks: \textit{design or provide feedback on research methodology or experiments, implement methods, assist with translation, and support qualitative and thematic data analysis}.
We have not used generative AI tools for the following tasks: \textit{help develop theoretical models or conceptual frameworks, formulate mathematical claims, provide critical ingredients for proving mathematical claims, propose or refine hypotheses, clean and reformat datasets, interpret results},
and the tasks \textit{generate synthetic data sets} and \textit{assist in the writing of proofs} are not applicable to this work.
Additionally, we used generative AI tools for the following tasks: \textit{create or modify scientific figures or images, create or edit software code, draft parts of a research paper, and edit a research paper to improve readability}. 
We have reviewed all AI-assisted work. 
Two authors verified and tested LLM-generated code and writing for correctness. 
We take responsibility for the final content,
including text, claims, or artifacts produced with generative AI.


\subsection*{Ethics statement}

Our method optimizes only the agent harness and does not update the underlying reasoning or image generation models.
Thus, it does not introduce new behaviors through model-weight modification, although it can change how existing model capabilities are composed.
As with image generation more broadly, improved multi-reference image generation may be misused to create misleading or unauthorized synthetic content.
The harness also inherits limitations and biases from the underlying models and evaluators.
We therefore encourage responsible use of the released code and harness.

\subsection*{Reproducibility statement}

We release the AutoRef implementation and AutoRef-Harness at an 
URL (\url{https://github.com/KuOnoda/AutoRef}),
together with the data splits and the pinned model revisions.
We conduct all evaluations on publicly available benchmarks.
The main paper describes the harness optimization procedure, model configurations, data splits, and evaluation protocol, while the Appendix provides additional evaluation details and complete results.


\subsubsection*{Acknowledgements}
We thank Google Japan for its funding support. 
MS was supported by JSPS KAKENHI Grant Number JP23H04974.

\bibliography{iclr2027_conference,related_work}

@inproceedings{hao2023promptist,
 author = {Hao, Yaru and Chi, Zewen and Dong, Li and Wei, Furu},
 booktitle = {Advances in Neural Information Processing Systems},
 doi = {10.52202/075280-2923},
 editor = {A. Oh and T. Naumann and A. Globerson and K. Saenko and M. Hardt and S. Levine},
 pages = {66923--66939},
 publisher = {Curran Associates, Inc.},
 title = {Optimizing Prompts for Text-to-Image Generation},
 url = {https://proceedings.neurips.cc/paper_files/paper/2023/file/d346d91999074dd8d6073d4c3b13733b-Paper-Conference.pdf},
 volume = {36},
 year = {2023}
}

@inproceedings{datta2024prompt_expansion,
    title = "Prompt Expansion for Adaptive Text-to-Image Generation",
    author = "Datta, Siddhartha  and
      Ku, Alexander  and
      Ramachandran, Deepak  and
      Anderson, Peter",
    booktitle = "Proceedings of the 62nd Annual Meeting of the Association for Computational Linguistics (Volume 1: Long Papers)",
    month = aug,
    year = "2024",
    address = "Bangkok, Thailand",
    publisher = "Association for Computational Linguistics",
    url = "https://aclanthology.org/2024.acl-long.189/",
    doi = "10.18653/v1/2024.acl-long.189",
    pages = "3449--3476",
}

@inproceedings{shen2023hugginggpt,
 author = {Shen, Yongliang and Song, Kaitao and Tan, Xu and Li, Dongsheng and Lu, Weiming and Zhuang, Yueting},
 booktitle = {Advances in Neural Information Processing Systems},
 doi = {10.52202/075280-1657},
 editor = {A. Oh and T. Naumann and A. Globerson and K. Saenko and M. Hardt and S. Levine},
 pages = {38154--38180},
 publisher = {Curran Associates, Inc.},
 title = {HuggingGPT: Solving AI Tasks with ChatGPT and its Friends in Hugging Face},
 url = {https://proceedings.neurips.cc/paper_files/paper/2023/file/77c33e6a367922d003ff102ffb92b658-Paper-Conference.pdf},
 volume = {36},
 year = {2023}
}

@inproceedings{wang2024genartist,
 author = {Wang, Zhenyu and Li, Aoxue and Li, Zhenguo and Liu, Xihui},
 booktitle = {Advances in Neural Information Processing Systems},
 doi = {10.52202/079017-4077},
 editor = {A. Globerson and L. Mackey and D. Belgrave and A. Fan and U. Paquet and J. Tomczak and C. Zhang},
 pages = {128374--128395},
 publisher = {Curran Associates, Inc.},
 title = {GenArtist: Multimodal LLM as an Agent for Unified Image Generation and Editing},
 url = {https://proceedings.neurips.cc/paper_files/paper/2024/file/e7c786024ca718f2487712bfe9f51030-Paper-Conference.pdf},
 volume = {37},
 year = {2024}
}

@inproceedings{guo2025comfymind,
 author = {Guo, Litao and Xu, Xinli and Wang, Luozhou and Lin, Jiantao and Zhou, Jinsong and Zhang, Zixin and Su, Bolan and Chen, Yingcong},
 booktitle = {Advances in Neural Information Processing Systems},
 doi = {10.52202/085713-1503},
 editor = {D. Belgrave and C. Zhang and H. Lin and R. Pascanu and P. Koniusz and M. Ghassemi and N. Chen},
 pages = {45128--45164},
 publisher = {Curran Associates, Inc.},
 title = {ComfyMind: Toward General-Purpose Generation via Tree-Based Planning and Reactive Feedback},
 url = {https://proceedings.neurips.cc/paper_files/paper/2025/file/40168e00bf87869c5d153e934d8a3602-Paper-Conference.pdf},
 volume = {38, Main Conference},
 year = {2025}
}

@inproceedings{yang2025idea2img,
  title={Idea2img: Iterative self-refinement with gpt-4v for automatic image design and generation},
  author={Yang, Zhengyuan and Wang, Jianfeng and Li, Linjie and Lin, Kevin and Lin, Chung-Ching and Liu, Zicheng and Wang, Lijuan},
  booktitle={European conference on computer vision},
  pages={167--184},
  year={2024},
  organization={Springer}
}

@article{wan2025maestro,
  title={Maestro: Self-improving text-to-image generation via agent orchestration},
  author={Wan, Xingchen and Zhou, Han and Sun, Ruoxi and Nakhost, Hootan and Jiang, Ke and Sinha, Rajarishi and Ar{\i}k, Sercan {\"O}},
  journal={arXiv preprint arXiv:2509.10704},
  year={2025}
}

@article{kovalev2025craft,
  title={{CRAFT}: Continuous reasoning and agentic feedback tuning for multimodal text-to-image generation},
  author={Kovalev, V and Kuvshinov, A and Buzovkin, A and Pokidov, D and Timonin, D},
  journal={arXiv preprint arXiv:2512.20362},
  year={2025}
}

@article{he2026gems,
  title={{GEMS}: Agent-native multimodal generation with memory and skills},
  author={He, Zefeng and Huang, Siyuan and Qu, Xiaoye and Li, Yafu and Zhu, Tong and Cheng, Yu and Yang, Yang},
  journal={arXiv preprint arXiv:2603.28088},
  year={2026}
}

@inproceedings{yao2026photoagent,
  title = {{PhotoAgent: Exploratory Visual Aesthetic Planning with Large Vision Models}},
  author = {Yao, Mingde and You, Zhiyuan and Tam, King-Man and Wang, Menglu and Xue, Tianfan},
  url = {https://icml.cc/virtual/2026/poster/63474},
  year = {2026},
  booktitle = {International Conference on Machine Learning},
}

@misc{meta2026muse,
  author       = {{Meta}},
  title        = {Introducing Muse Image: Image Generation Built for Your World},
  year         = {2026},
  month        = jul,
  howpublished = {Meta Newsroom},
  url          = {https://about.fb.com/news/2026/07/introducing-muse-image-meta-ai/},
  note         = {Accessed: 2026-09-15}
}

@article{jiang2026genagent,
  title={Genagent: Scaling text-to-image generation via agentic multimodal reasoning},
  author={Jiang, Kaixun and Wang, Yuzheng and Zhou, Junjie and Li, Pandeng and Liu, Zhihang and Xie, Chen-Wei and Chen, Zhaoyu and Zheng, Yun and Zhang, Wenqiang},
  journal={The 19th European Conference on Computer Vision},
  year={2026}
}

@article{chen2026genevolve,
  title={GenEvolve: Self-Evolving Image Generation Agents via Tool-Orchestrated Visual Experience Distillation},
  author={Chen, Sixiang and Xing, Zhaohu and Ye, Tian and Geng, Xinyu and Lin, Yunlong and Lai, Jianyu and He, Xuanhua and Zhai, Fuxiang and Gao, Jialin and Zhu, Lei},
  journal={arXiv preprint arXiv:2605.21605},
  year={2026}
}

@article{zhao2026toolartist,
  title={ToolArtist: Tool-Using Unified Multimodal Models for Agentic Image Generation},
  author={Zhao, Jiahao and Yu, Xiaomin and Sun, Zhongxiang and Teng, Fengwei and Qin, Chengwei and Hu, Xiaobin and Xu, Jun and Yan, Shuicheng},
  journal={arXiv preprint arXiv:2608.04436},
  year={2026}
}

@inproceedings{zhou2023ape,
  title = {{Large Language Models Are Human-Level Prompt Engineers}},
  author={Yongchao Zhou and Andrei Ioan Muresanu and Ziwen Han and Keiran Paster and Silviu Pitis and Harris Chan and Jimmy Ba},
  url = {https://openreview.net/forum?id=92gvk82DE-},
  year = {2023},
  booktitle = {International Conference on Learning Representations},
}

@inproceedings{yang2024opro,
 author = {Yang, Chengrun and Wang, Xuezhi and Lu, Yifeng and Liu, Hanxiao and Le, Quoc V and Zhou, Denny and Chen, Xinyun},
 booktitle = {International Conference on Learning Representations},
 editor = {B. Kim and Y. Yue and S. Chaudhuri and K. Fragkiadaki and M. Khan and Y. Sun},
 pages = {12028--12068},
 title = {Large Language Models as Optimizers},
 url = {https://proceedings.iclr.cc/paper_files/paper/2024/file/3339f19c5fcee3ad74502947a32be9e6-Paper-Conference.pdf},
 volume = {2024},
 year = {2024}
}

@inproceedings{pryzant2023protegi,
    title = "Automatic Prompt Optimization with ``Gradient Descent'' and Beam Search",
    author = "Pryzant, Reid  and
      Iter, Dan  and
      Li, Jerry  and
      Lee, Yin  and
      Zhu, Chenguang  and
      Zeng, Michael",
    editor = "Bouamor, Houda  and
      Pino, Juan  and
      Bali, Kalika",
    booktitle = "Proceedings of the 2023 Conference on Empirical Methods in Natural Language Processing",
    month = dec,
    year = "2023",
    address = "Singapore",
    publisher = "Association for Computational Linguistics",
    url = "https://aclanthology.org/2023.emnlp-main.494/",
    doi = "10.18653/v1/2023.emnlp-main.494",
    pages = "7957--7968"
}

@inproceedings{guo2024evoprompt,
 author = {Guo, Qingyan and Wang, Rui and Guo, Junliang and Li, Bei and Song, Kaitao and Tan, Xu and Liu, Guoqing and Bian, Jiang and Yang, Yujiu},
 booktitle = {International Conference on Learning Representations},
 editor = {B. Kim and Y. Yue and S. Chaudhuri and K. Fragkiadaki and M. Khan and Y. Sun},
 pages = {34133--34156},
 title = {Connecting Large Language Models with Evolutionary Algorithms Yields Powerful Prompt Optimizers},
 url = {https://proceedings.iclr.cc/paper_files/paper/2024/file/9156b0f6dfa9bbd18c79cc459ef5d61c-Paper-Conference.pdf},
 volume = {2024},
 year = {2024}
}

@inproceedings{khattab2024dspy,
 author = {Khattab, Omar and Singhvi, Arnav and Maheshwari, Paridhi and Zhang, Zhiyuan and Santhanam, Keshav and A, Sri Vardhamanan and Haq, Saiful and Sharma, Ashutosh and Joshi, Thomas and Moazam, Hanna and Miller, Heather and Zaharia, Matei and Potts, Christopher},
 booktitle = {International Conference on Learning Representations},
 editor = {B. Kim and Y. Yue and S. Chaudhuri and K. Fragkiadaki and M. Khan and Y. Sun},
 pages = {54928--54958},
 title = {DSPy: Compiling Declarative Language Model Calls into State-of-the-Art Pipelines},
 url = {https://proceedings.iclr.cc/paper_files/paper/2024/file/f1cf02ce09757f57c3b93c0db83181e0-Paper-Conference.pdf},
 volume = {2024},
 year = {2024}
}

@inproceedings{opsahlong2024mipro,
    title = "Optimizing Instructions and Demonstrations for Multi-Stage Language Model Programs",
    author = "Opsahl-Ong, Krista  and
      Ryan, Michael J  and
      Purtell, Josh  and
      Broman, David  and
      Potts, Christopher  and
      Zaharia, Matei  and
      Khattab, Omar",
    editor = "Al-Onaizan, Yaser  and
      Bansal, Mohit  and
      Chen, Yun-Nung",
    booktitle = "Proceedings of the 2024 Conference on Empirical Methods in Natural Language Processing",
    month = nov,
    year = "2024",
    address = "Miami, Florida, USA",
    publisher = "Association for Computational Linguistics",
    url = "https://aclanthology.org/2024.emnlp-main.525/",
    doi = "10.18653/v1/2024.emnlp-main.525",
    pages = "9340--9366"
}

@article{yuksekgonul2025textgrad,
  title = {{Optimizing generative AI by backpropagating language model feedback}},
  author = {Yuksekgonul, Mert and Bianchi, Federico and Boen, Joseph and Liu, Sheng and Lu, Pan and Huang, Zhi and Guestrin, Carlos and Zou, James},
  url = {https://www.nature.com/articles/s41586-025-08661-4},
  year = {2025},
  journal = {Nature},
  volume = {639},
  number = {8055},
  pages = {609--616},
  doi = {10.1038/s41586-025-08661-4},
}

@inproceedings{agrawal2026gepa,
 author = {Agrawal, Lakshya A and Tan, Shangyin and Soylu, Dilara and Ziems, Noah and Khare, Rishi and Opsahl-Ong, Krista and Singhvi, Arnav and Shandilya, Herumb and Ryan, Michael J and Jiang, Meng and Potts, Christopher and Sen, Koushik and Dimakis, Alex and Stoica, Ion and Klein, Dan and Zaharia, Matei and Khattab, Omar},
 booktitle = {International Conference on Learning Representations},
 editor = {C. Vondrick and B. Hariharan and C. Raffel and L. Pinto and D. Yang and A. Faust},
 pages = {8479--8565},
 title = {GEPA: Reflective Prompt Evolution Can Outperform Reinforcement Learning},
 url = {https://proceedings.iclr.cc/paper_files/paper/2026/file/0e9e708b6f48e14fd0ac29e167413f76-Paper-Conference.pdf},
 volume = {2026},
 year = {2026}
}

@InProceedings{zhuge2024gptswarm,
  title = 	 {{GPTS}warm: Language Agents as Optimizable Graphs},
  author =       {Zhuge, Mingchen and Wang, Wenyi and Kirsch, Louis and Faccio, Francesco and Khizbullin, Dmitrii and Schmidhuber, J\"{u}rgen},
  booktitle = 	 {Proceedings of the 41st International Conference on Machine Learning},
  pages = 	 {62743--62767},
  year = 	 {2024},
  editor = 	 {Salakhutdinov, Ruslan and Kolter, Zico and Heller, Katherine and Weller, Adrian and Oliver, Nuria and Scarlett, Jonathan and Berkenkamp, Felix},
  volume = 	 {235},
  series = 	 {Proceedings of Machine Learning Research},
  month = 	 {21--27 Jul},
  publisher =    {PMLR},
  url = 	 {https://proceedings.mlr.press/v235/zhuge24a.html}
}

@inproceedings{hu2025adas,
 author = {Hu, Shengran and Lu, Cong and Clune, Jeff},
 booktitle = {International Conference on Learning Representations},
 editor = {Y. Yue and A. Garg and N. Peng and F. Sha and R. Yu},
 pages = {21344--21377},
 title = {Automated Design of Agentic Systems},
 url = {https://proceedings.iclr.cc/paper_files/paper/2025/file/36b7acf6f6010652b3f2a433774a66fe-Paper-Conference.pdf},
 volume = {2025},
 year = {2025}
}

@inproceedings{zhang2025aflow,
 author = {Zhang, Jiayi and Xiang, Jinyu and Yu, Zhaoyang and Teng, Fengwei and Chen, XiongHui and Chen, Jiaqi and Zhuge, Mingchen and Cheng, Xin and Hong, Sirui and Wang, Jinlin and Zheng, Bingnan and Liu, Bang and Luo, Yuyu and Wu, Chenglin},
 booktitle = {International Conference on Learning Representations},
 editor = {Y. Yue and A. Garg and N. Peng and F. Sha and R. Yu},
 pages = {34040--34077},
 title = {AFlow: Automating Agentic Workflow Generation},
 url = {https://proceedings.iclr.cc/paper_files/paper/2025/file/5492ecbce4439401798dcd2c90be94cd-Paper-Conference.pdf},
 volume = {2025},
 year = {2025}
}

@article{novikov2025alphaevolve,
  title={{AlphaEvolve}: A coding agent for scientific and algorithmic discovery},
  author={Novikov, Alexander and V{\~u}, Ng{\^a}n and Eisenberger, Marvin and Dupont, Emilien and Huang, Po-Sen and Wagner, Adam Zsolt and Shirobokov, Sergey and Kozlovskii, Borislav and Ruiz, Francisco JR and Mehrabian, Abbas and others},
  journal={arXiv preprint arXiv:2506.13131},
  year={2025}
}

@inproceedings{lange2026shinkaevolve,
 author = {Lange, Robert and Imajuku, Yuki and Cetin, Edoardo},
 booktitle = {International Conference on Learning Representations},
 editor = {C. Vondrick and B. Hariharan and C. Raffel and L. Pinto and D. Yang and A. Faust},
 pages = {74026--74078},
 title = {ShinkaEvolve: Towards Open-Ended and Sample-Efficient Program Evolution},
 url = {https://proceedings.iclr.cc/paper_files/paper/2026/file/7886b9bafe76c52fd568db10ff9772df-Paper-Conference.pdf},
 volume = {2026},
 year = {2026}
}

@inproceedings{zelikman2024stop,
  title = {{Self-Taught Optimizer (STOP): Recursively Self-Improving Code Generation}},
  author=  {Zelikman, Eric and Lorch, Eliana and Mackey, Lester and Kalai, Adam Tauman},
  url = {https://openreview.net/forum?id=46Zgqo4QIU},
  year = {2024},
  booktitle = {Conference on Language Modeling},
}

@inproceedings{robeyns2025sica,
  title = {{A Self-Improving Coding Agent}},
  author = {Robeyns, Maxime and Szummer, Martin and Aitchison, Laurence},
  url = {https://openreview.net/forum?id=rShJCyLsOr},
  year = {2025},
  booktitle = {ICLR 2025 Workshop on Scaling Self-Improving Foundation Models},
}

@inproceedings{zhang2026dgm,
 author = {Zhang, Jenny and Hu, Shengran and Lu, Cong and Lange, Robert and Clune, Jeff},
 booktitle = {International Conference on Learning Representations},
 editor = {C. Vondrick and B. Hariharan and C. Raffel and L. Pinto and D. Yang and A. Faust},
 pages = {104223--104294},
 title = {Darwin G\"{o}del Machine: Open-Ended Evolution of Self-Improving Agents},
 url = {https://proceedings.iclr.cc/paper_files/paper/2026/file/aa5f5e6eb6f613ec412f1d948dfa21a5-Paper-Conference.pdf},
 volume = {2026},
 year = {2026}
}

@inproceedings{
lee2026metaharness,
title={Meta-Harness: End-to-End Optimization of Model Harnesses},
author={Yoonho Lee and Roshen Sanjay Nair and Qizheng Zhang and Kangwook Lee and Omar Khattab and Chelsea Finn},
booktitle={Third Conference on Language Modeling},
year={2026},
url={https://openreview.net/forum?id=tmbOUyFx3R}
}

@article{luo2026autodesign,
  title={AutoDesign: Meta-Harness Optimization for Long-Horizon Agentic Design},
  author={Luo, Yaxin and Jiang, Haobin and Zou, Jialv and Huang, Xu and Yan, Wenhao and Li, Haodong and Yue, Zhengrong and Li, Jing and Chen, Xiaofu and Zhao, Xiaohan and others},
  journal={arXiv preprint arXiv:2608.13560},
  year={2026}
}

@String(CVPR= {IEEE Conf. Comput. Vis. Pattern Recog.})

@String(ECCV= {Eur. Conf. Comput. Vis.})

@String(ICLR = {Int. Conf. Learn. Represent.})

@String(AAAI = {AAAI})

@String(CVPR  = {CVPR})

@String(ECCV  = {ECCV})

@String(ICLR  = {ICLR})

@inproceedings{
yeh2024sampling,
title={Training-Free Diffusion Model Alignment with Sampling Demons},
author={Po-Hung Yeh and Kuang-Huei Lee and Jun-cheng Chen},
booktitle={The Thirteenth International Conference on Learning Representations},
year={2025},
url={https://openreview.net/forum?id=tfemquulED}
}

@article{furuta2024improving,
      title={Improving Dynamic Object Interactions in Text-to-Video Generation with AI Feedback}, 
      author={Hiroki Furuta and Heiga Zen and Dale Schuurmans and Aleksandra Faust and Yutaka Matsuo and Percy Liang and Sherry Yang},
      year={2024},
      journal={arXiv preprint arXiv:2412.02617},
}

@inproceedings{
snell2024scaling,
title={Scaling {LLM} Test-Time Compute Optimally Can be More Effective than Scaling Parameters for Reasoning},
author={Charlie Victor Snell and Jaehoon Lee and Kelvin Xu and Aviral Kumar},
booktitle={The Thirteenth International Conference on Learning Representations},
year={2025},
url={https://openreview.net/forum?id=4FWAwZtd2n}
}

@article{openai2023gpt4,
      title={GPT-4 Technical Report}, 
      author={OpenAI},
      year={2023},
      journal={arXiv preprint arXiv:2303.08774},
}

@inproceedings{
singhal2025general,
title={A General Framework for Inference-time Scaling and Steering of Diffusion Models},
author={Raghav Singhal and Zachary Horvitz and Ryan Teehan and Mengye Ren and Zhou Yu and Kathleen McKeown and Rajesh Ranganath},
booktitle={Forty-second International Conference on Machine Learning},
year={2025},
url={https://openreview.net/forum?id=Jp988ELppQ}
}

@inproceedings{ma2025inferencetime,
    author    = {Ma, Nanye and Tong, Shangyuan and Jia, Haolin and Hu, Hexiang and Su, Yu-Chuan and Zhang, Mingda and Yang, Xuan and Li, Yandong and Jaakkola, Tommi and Jia, Xuhui and Xie, Saining},
    title     = {Scaling Inference Time Compute for Diffusion Models},
    booktitle = {Proceedings of the IEEE/CVF Conference on Computer Vision and Pattern Recognition (CVPR)},
    month     = {June},
    year      = {2025},
    pages     = {2523-2534}
}

@inproceedings{kim2025das,
      title={Test-time Alignment of Diffusion Models without Reward Over-optimization}, 
      author={Sunwoo Kim and Minkyu Kim and Dongmin Park},
      year={2025}, 
      booktitle={The Thirteenth International Conference on Learning Representations},
}

@inproceedings{
    oshima2025inference,
    title={Inference-Time Text-to-Video Alignment with Diffusion Latent Beam Search},
    author={Yuta Oshima and Masahiro Suzuki and Yutaka Matsuo and Hiroki Furuta},
    booktitle={The Thirty-ninth Annual Conference on Neural Information Processing Systems},
    year={2025},
    url={https://openreview.net/forum?id=c9EAmyYPOv}
}

@misc{google2025nanobanana,
  author       = {{Google DeepMind}},
  title        = {Nano Banana: Gemini 2.5 Flash Image Model},
  year         = {2025},
  howpublished = {\url{https://developers.googleblog.com/en/introducing-gemini-2-5-flash-image/}},
  note         = {Accessed: 2025-10-31},
  institution  = {Google DeepMind}
}

@misc{google2025nanobananapro,
  author       = {{Google DeepMind}},
  title        = {Nano Banana Pro},
  year         = {2025},
  howpublished = {\url{https://deepmind.google/models/gemini-image/pro/}},
  note         = {Accessed: 2025-11-27},
  institution  = {Google DeepMind}
}

@misc{openai2025gpt4oimage,
  author       = {{OpenAI}},
  title        = {GPT-4o Image Generation},
  year         = {2025},
  howpublished = {\url{https://openai.com/index/introducing-4o-image-generation/}},
  note         = {Accessed: 2025-10-31},
  institution  = {OpenAI}
}

@inproceedings{zhu2023tryondiffusion,
  author={Zhu, Luyang and Yang, Dawei and Zhu, Tyler and Reda, Fitsum and Chan, William and Saharia, Chitwan and Norouzi, Mohammad and Kemelmacher-Shlizerman, Ira},
  title={TryOnDiffusion: A Tale of Two UNets},
  booktitle = {Proceedings of the IEEE/CVF Conference on Computer Vision and Pattern Recognition (CVPR)},
  month = {June},
  year={2023},
  pages = {4606-4615}
}

@InProceedings{ruiz2022dreambooth,
    author    = {Ruiz, Nataniel and Li, Yuanzhen and Jampani, Varun and Pritch, Yael and Rubinstein, Michael and Aberman, Kfir},
    title     = {DreamBooth: Fine Tuning Text-to-Image Diffusion Models for Subject-Driven Generation},
    booktitle = {Proceedings of the IEEE/CVF Conference on Computer Vision and Pattern Recognition (CVPR)},
    month     = {June},
    year      = {2023},
    pages     = {22500-22510}
}

@inproceedings{wu2025omnigen2,
  title={Omnigen2: Towards instruction-aligned multimodal generation},
  author={Wu, Chenyuan and Wang, Jiahao and Zheng, Pengfei and Yan, Ruiran and Xiao, Shitao and Luo, Xin and Wang, Yueze and Li, Wanli and Jiang, Xiyan and Liu, Yexin and others},
  booktitle={Proceedings of the IEEE/CVF Conference on Computer Vision and Pattern Recognition},
  pages={21964--21975},
  year={2026}
}

@inproceedings{xia2025dreamomni2,
    author    = {Xia, Bin and peng, bohao and Zhang, Yuechen and Huang, Junjia and Liu, Jiyang and Li, Jingyao and Tan, Haoru and Wu, Sitong and Wang, Chengyao and Wang, Yitong and Yu, Bei and Jia, Jiaya},
    title     = {DreamOmni2: Multimodal Instruction-based Generation and Editing},
    booktitle = {Proceedings of the IEEE/CVF Conference on Computer Vision and Pattern Recognition (CVPR)},
    month     = {June},
    year      = {2026},
    pages     = {29275-29284}
}

@misc{wu2025qwenimagetechnicalreport,
      title={Qwen-Image Technical Report}, 
      author={Chenfei Wu and Jiahao Li and Jingren Zhou and Junyang Lin and Kaiyuan Gao and Kun Yan and Sheng-ming Yin and Shuai Bai and Xiao Xu and Yilei Chen and Yuxiang Chen and Zecheng Tang and Zekai Zhang and Zhengyi Wang and An Yang and Bowen Yu and Chen Cheng and Dayiheng Liu and Deqing Li and Hang Zhang and Hao Meng and Hu Wei and Jingyuan Ni and Kai Chen and Kuan Cao and Liang Peng and Lin Qu and Minggang Wu and Peng Wang and Shuting Yu and Tingkun Wen and Wensen Feng and Xiaoxiao Xu and Yi Wang and Yichang Zhang and Yongqiang Zhu and Yujia Wu and Yuxuan Cai and Zenan Liu},
      year={2025},
      eprint={2508.02324},
      archivePrefix={arXiv},
      primaryClass={cs.CV},
      url={https://arxiv.org/abs/2508.02324}, 
}

@inproceedings{morita2025tkg,
    author    = {Morita, Ryugo and Frolov, Stanislav and Moser, Brian Bernhard and Shirakawa, Takahiro and Watanabe, Ko and Dengel, Andreas and Zhou, Jinjia},
    title     = {TKG-DM: Training-free Chroma Key Content Generation Diffusion Model},
    booktitle = {Proceedings of the IEEE/CVF Conference on Computer Vision and Pattern Recognition (CVPR)},
    month     = {June},
    year      = {2025},
    pages     = {13031-13040}
}

@misc{bai2025qwen3vl,
      title={Qwen3-VL Technical Report}, 
      author={Shuai Bai and Yuxuan Cai and Ruizhe Chen and Keqin Chen and Xionghui Chen and Zesen Cheng and Lianghao Deng and Wei Ding and Chang Gao and Chunjiang Ge and Wenbin Ge and Zhifang Guo and Qidong Huang and Jie Huang and Fei Huang and Binyuan Hui and Shutong Jiang and Zhaohai Li and Mingsheng Li and Mei Li and Kaixin Li and Zicheng Lin and Junyang Lin and Xuejing Liu and Jiawei Liu and Chenglong Liu and Yang Liu and Dayiheng Liu and Shixuan Liu and Dunjie Lu and Ruilin Luo and Chenxu Lv and Rui Men and Lingchen Meng and Xuancheng Ren and Xingzhang Ren and Sibo Song and Yuchong Sun and Jun Tang and Jianhong Tu and Jianqiang Wan and Peng Wang and Pengfei Wang and Qiuyue Wang and Yuxuan Wang and Tianbao Xie and Yiheng Xu and Haiyang Xu and Jin Xu and Zhibo Yang and Mingkun Yang and Jianxin Yang and An Yang and Bowen Yu and Fei Zhang and Hang Zhang and Xi Zhang and Bo Zheng and Humen Zhong and Jingren Zhou and Fan Zhou and Jing Zhou and Yuanzhi Zhu and Ke Zhu},
      year={2025},
      eprint={2511.21631},
      archivePrefix={arXiv},
      primaryClass={cs.CV},
      url={https://arxiv.org/abs/2511.21631}, 
}

@inproceedings{
chong2024catvton,
title={Cat{VTON}: Concatenation Is All You Need for Virtual Try-On with Diffusion Models},
author={Zheng Chong and Xiao Dong and Haoxiang Li and shiyue Zhang and Wenqing Zhang and Hanqing Zhao and xujie zhang and Dongmei Jiang and Xiaodan Liang},
booktitle={The Thirteenth International Conference on Learning Representations},
year={2025},
url={https://openreview.net/forum?id=jt1h2dnmng}
}

@inproceedings{inoue2023layout,
    title={{LayoutDM: Discrete Diffusion Model for Controllable Layout Generation}},
    author={Naoto Inoue and Kotaro Kikuchi and Edgar Simo-Serra and Mayu Otani and Kota Yamaguchi},
    booktitle={Proceedings of the IEEE/CVF Conference on Computer Vision and Pattern Recognition (CVPR)},
    year={2023},
    pages={10167-10176},
  }

@inproceedings{oshima2026multibanana,
    author    = {Oshima, Yuta and Miyake, Daiki and Matsutani, Kohsei and Iwasawa, Yusuke and Suzuki, Masahiro and Matsuo, Yutaka and Furuta, Hiroki},
    title     = {MultiBanana: A Challenging Benchmark for Multi-Reference Text-to-Image Generation},
    booktitle = {Proceedings of the IEEE/CVF Conference on Computer Vision and Pattern Recognition (CVPR)},
    month     = {June},
    year      = {2026},
    pages     = {448-460}
}

@inproceedings{
    xu2026contextgen,
    title={ContextGen: Contextual Layout Anchoring for Identity-Consistent Multi-Instance Generation},
    author={Ruihang Xu and Dewei Zhou and Fan Ma and Yi Yang},
    booktitle={The Fourteenth International Conference on Learning Representations},
    year={2026}
}

@inproceedings{cvpr2026garments2look,
    title={Garments2Look: A Multi-Reference Dataset for High-Fidelity Outfit-Level Virtual Try-On with Clothing and Accessories},
    author={Hu, Junyao and Cheng, Zhongwei and Wong, Waikeung and Zou, Xingxing},
    booktitle={Proceedings of the IEEE/CVF Conference on Computer Vision and Pattern Recognition (CVPR)},
    year={2026}
}

@misc{google2025nanobanana2,
  author       = {{Google DeepMind}},
  title        = {Nano Banana 2: Combining Pro capabilities with lightning-fast speed},
  year         = {2026},
  howpublished = {\url{https://blog.google/innovation-and-ai/technology/ai/nano-banana-2/}},
  note         = {Accessed: 2026-4-30},
  institution  = {Google DeepMind}
}

@misc{openai2025gptimage1_5,
  author       = {{OpenAI}},
  title        = {The new ChatGPT Images is here},
  year         = {2025},
  howpublished = {\url{https://openai.com/index/new-chatgpt-images-is-here/}},
  note         = {Accessed: 2026-4-30},
  institution  = {OpenAI}
}

@article{deng2025bagel,
  title   = {Emerging Properties in Unified Multimodal Pretraining},
  author  = {Deng, Chaorui and Zhu, Deyao and Li, Kunchang and Gou, Chenhui and Li, Feng and Wang, Zeyu and Zhong, Shu and Yu, Weihao and Nie, Xiaonan and Song, Ziang and Shi, Guang and Fan, Haoqi},
  journal = {arXiv preprint arXiv:2505.14683},
  year    = {2025}
}

@inproceedings{zhang2026rcedit,
  title={RCEdit-500K: Reference Completion for Image-Conditioned Image Editing},
  author={Zhang, Jingxu and Kim, Daneul and Pan, Yueming and Chen, Dong and Qiu, Kai and Liu, Yang and Yang, Yifan and Dai, Qi and Sun, Xiaoyan and Luo, Chong},
  booktitle={European Conference on Computer Vision (ECCV)},
  year={2026}
}

@inproceedings{huang2026scaling,
      title={Scaling Multi-Reference Image Generation with Dynamic Reward Optimization}, 
      author={Wenwang Huang and Yusen Fu and Junjie Wang and Mengfei Huang and Yulin Li and Gan Liu and Jing Cai and Yancheng He and Zhuotao Tian},
      year={2026},
      booktitle = {The 19th European Conference on Computer Vision} 
}

@misc{
he2026evosearch,
title={Scaling Image and Video Generation via Test-Time Evolutionary Search},
author={Haoran He and Jiajun Liang and Xintao Wang and Pengfei Wan and Kun Gai and Ling Pan},
year={2026},
url={https://openreview.net/forum?id=CFlOUNWsaP}
}

@inproceedings{kojima2022large,
 author = {Kojima, Takeshi and Gu, Shixiang (Shane) and Reid, Machel and Matsuo, Yutaka and Iwasawa, Yusuke},
 booktitle = {Advances in Neural Information Processing Systems},
 pages = {22199--22213},
 title = {Large Language Models are Zero-Shot Reasoners},
 volume = {35},
 year = {2022}
}

@inproceedings{qu2026scalespeed,
    author    = {Qu, Xiangyan and Yuan, Zhenlong and Tang, Jing and Chen, Rui and Tang, Datao and Yu, Meng and Sun, Lei and Bai, Yancheng and Chu, Xiangxiang and Gou, Gaopeng and Xiong, Gang and Cai, Yujun},
    title     = {From Scale to Speed: Adaptive Test-Time Scaling for Image Editing},
    booktitle = {Proceedings of the IEEE/CVF Conference on Computer Vision and Pattern Recognition (CVPR)},
    month     = {June},
    year      = {2026},
    pages     = {23272-23282}
}

@article{wu2025imagerysearch,
    title={ImagerySearch: Adaptive Test-Time Search for Video Generation Beyond Semantic Dependency Constraints}, 
    author={Meiqi Wu and Jiashu Zhu and Xiaokun Feng and Chubin Chen and Chen Zhu and Bingze Song and Fangyuan Mao and Jiahong Wu and Xiangxiang Chu and Kaiqi Huang},
    journal = {Proceedings of the AAAI Conference on Artificial Intelligence},
    volume  = {40},
    number  = {13},
    pages   = {10700--10708},
    year    = {2026},
    month   = {Mar.},
    doi     = {10.1609/aaai.v40i13.38044},
    url     = {https://ojs.aaai.org/index.php/AAAI/article/view/38044}
}

@inproceedings{zhao2026latsearch,
  title   = {LatSearch: Latent Reward-Guided Search for Faster Inference-Time Scaling in Video Diffusion},
  author  = {Zhao, Zengqun and Liu, Ziquan and Cao, Yu and Gong, Shaogang and Zhang, Zhensong and Song, Jifei and Deng, Jiankang and Patras, Ioannis},
  booktitle = {European Conference on Computer Vision (ECCV)},
  year    = {2026}
}

@inproceedings{rawal2026flashbon,
      title={Flash-BoN: Instant Drafts for Inference-Time Scaling in Diffusion Models}, 
      author={Ruchit Rawal and Reza Shirkavand and Sayak Paul and Yuxin Wen and Heng Huang and Yizheng Chen and Tom Goldstein and Gowthami Somepalli},
      year={2026},
      booktitle={The 19th European Conference on Computer Vision}
}

@article{
jung2026inferencetime,
title={Inference-Time Scaling for Joint Audio-Video Generation},
author={Jaemin Jung and Kyeongha Rho and Inkyu Shin and Joon Son Chung},
journal={Transactions on Machine Learning Research},
issn={2835-8856},
year={2026},
url={https://openreview.net/forum?id=MHNFjjm5nO}
}

@misc{saini2026cachedsearch,
      title={CachedSearch: Training-Free Cached Exploration for Test-Time Search in Video Diffusion}, 
      author={Shreshth Saini and Neil Birkbeck and Yilin Wang and Balu Adsumilli and Alan C. Bovik},
      year={2026},
      eprint={2607.23159},
      archivePrefix={arXiv},
      primaryClass={cs.AI},
      url={https://arxiv.org/abs/2607.23159}, 
}

@article{guimaraes2026psp,
  title={Inference-Time Scaling of Diffusion Models via Progressive Seed Pruning},
  author={Guimaraes, Rogerio and Perona, Pietro},
  journal={arXiv preprint arxiv:2607.21591},
  year={2026}
}

@inproceedings{
zhang2026classical,
title={Inference-time scaling of diffusion models through classical search},
author={XiangCheng Zhang and Haowei Lin and Haotian Ye and James Zou and Jianzhu Ma and Yitao Liang and Yilun Du},
booktitle={The Fourteenth International Conference on Learning Representations},
year={2026},
url={https://openreview.net/forum?id=b7Ftp6U78i}
}

@article{miyai2026taskcoevolve,
  title   = {Task-CoEvolve: Efficient Harness Optimization via Adaptive Validation Task Selection},
  author  = {Miyai, Atsuyuki and Aizawa, Kiyoharu and Yamasaki, Toshihiko},
  journal = {arXiv preprint arxiv:2608.20169},
  year    = {2026}
}

@misc{anthropic2025claudecode,
  author       = {{Anthropic}},
  title        = {Claude Code},
  year         = {2025},
  howpublished = {\url{https://claude.com/product/claude-code}},
  note         = {Accessed: 2026-09-18}
}

@misc{zhang2026selfharnes,
      title={Self-Harness: Harnesses That Improve Themselves},
      author={Hangfan Zhang and Shao Zhang and Kangcong Li and Chen Zhang and Yang Chen and Yiqun Zhang and Lei Bai and Shuyue Hu},
      year={2026},
      eprint={2606.09498},
      archivePrefix={arXiv},
      primaryClass={cs.CL},
      url={https://arxiv.org/abs/2606.09498},
}

@misc{lin2026agenticharness,
      title={Agentic Harness Engineering: Observability-Driven Automatic Evolution of Coding-Agent Harnesses}, 
      author={Jiahang Lin and Shichun Liu and Chengjun Pan and Lizhi Lin and Shihan Dou and Zhiheng Xi and Xuanjing Huang and Hang Yan and Zhenhua Han and Tao Gui and Yu-Gang Jiang},
      year={2026},
      eprint={2604.25850},
      archivePrefix={arXiv},
      primaryClass={cs.CL},
      url={https://arxiv.org/abs/2604.25850}, 
}

@misc{onoda2026multiaxis,
      title={Multi-Axis Max@K Reinforcement Learning for Representative Diversity in Text-to-Image Generation}, 
      author={Ku Onoda and Paavo Parmas and Hiroki Furuta and Soichiro Nishimori and Yuta Oshima and Shohei Taniguchi and Yutaka Matsuo},
      year={2026},
      eprint={2607.14962},
      archivePrefix={arXiv},
      primaryClass={cs.LG},
      url={https://arxiv.org/abs/2607.14962}, 
}

@inproceedings{
    carion2025sam3segmentconcepts,
    title={{SAM} 3: Segment Anything with Concepts},
    author={Nicolas Carion and Laura Gustafson and Yuan-Ting Hu and Shoubhik Debnath and Ronghang Hu and Didac Suris Coll-Vinent and Chaitanya Ryali and Kalyan Vasudev Alwala and Haitham Khedr and Andrew Huang and Jie Lei and Tengyu Ma and Baishan Guo and Arpit Kalla and Markus Marks and Joseph Greer and Meng Wang and Peize Sun and Roman R{\"a}dle and Triantafyllos Afouras and Effrosyni Mavroudi and Katherine Xu and Tsung-Han Wu and Yu Zhou and Liliane Momeni and RISHI HAZRA and Shuangrui Ding and Sagar Vaze and Francois Porcher and Feng Li and Siyuan Li and Aishwarya Kamath and Ho Kei Cheng and Piotr Dollar and Nikhila Ravi and Kate Saenko and Pengchuan Zhang and Christoph Feichtenhofer},
    booktitle={The Fourteenth International Conference on Learning Representations},
    year={2026},
    url={https://openreview.net/forum?id=r35clVtGzw}
}

@misc{huang2026envharness,
      title={EnvHarness: Awakening Static Worlds for Agent Learning}, 
      author={Chengsong Huang and Zifeng Wang and Rujun Han and Jun Yan and Yanfei Chen and Zoey CuiZhu and Ke Jiang and Peng Xia and Han Yu and Yufan Zhuang and Yifei Ming and Jiaqi Pan and Bhavana Dalvi Mishra and Jiaxin Huang and Burak Gokturk and Tomas Pfister and Chen-Yu Lee},
      year={2026},
      eprint={2608.19880},
      archivePrefix={arXiv},
      primaryClass={cs.AI},
      url={https://arxiv.org/abs/2608.19880}, 
}

@misc{bytedance2025seedream45,
  author       = {{ByteDance Seed}},
  title        = {Seedream 4.5},
  year         = {2025},
  howpublished = {\url{https://seed.bytedance.com/en/seedream4_5}},
  note         = {Accessed: 2026-09-23}
}

@misc{openai2026gpt55,
  author       = {{OpenAI}},
  title        = {Introducing {GPT-5.5}},
  year         = {2026},
  howpublished = {\url{https://openai.com/index/introducing-gpt-5-5/}},
  note         = {Accessed: 2026-09-24}
}

@misc{blackforestlabs2026flux2klein,
  author       = {{Black Forest Labs}},
  title        = {{FLUX.2 [klein]}},
  year         = {2026},
  howpublished = {\url{https://bfl.ai/models/flux-2-klein}},
  note         = {Accessed: 2026-09-24}
}
\bibliographystyle{iclr2027_conference}

\clearpage
\appendix
\section*{Appendix}
\makeatletter\setlength{\@fptop}{0pt}\setlength{\@fpsep}{12pt plus 1fil}\makeatother

\section{Implementation Details}
\label{app:impl}
The proposer is Claude Fable 5.1, run through the Claude Code CLI with the \texttt{Read}, \texttt{Glob}, \texttt{Grep}, \texttt{Write}, \texttt{Edit}, and \texttt{Bash} tools. It receives only text at the start of each session and loads images from the filesystem when it needs to inspect them. The search runs for five iterations, and the search and all evaluations run on NVIDIA RTX 6000 Ada GPUs.
\autoref{tab:model-versions} lists the models and their identifiers. The reasoning model is a dated API snapshot rather than a moving alias, open-weight models are pinned to fixed revisions in the released code, and the evaluator decodes greedily. 
The proprietary models are called through their APIs with all references and the instruction, a square output, and otherwise default settings: GPT-Image-1.5 through the image edit endpoint at $1024\times1024$, Nano Banana Pro through the Gemini API with a $1{:}1$ aspect ratio, and Seedream 4.5 at its native $2048\times2048$, downsampled to $1024\times1024$ before evaluation.
Code is available at \url{https://github.com/KuOnoda/AutoRef}.

\begin{table}[h]
\centering
\caption{Model versions used in this work. Identifiers are Hugging Face repositories or API model
names. Exact revisions are pinned in the released code.}
\label{tab:model-versions}
\small
\begin{tabular}{l l}
\toprule
Model & Identifier \\
\midrule
\multicolumn{2}{l}{\textit{Open image models}} \\
OmniGen2~\citep{wu2025omnigen2} & \texttt{\footnotesize OmniGen2/OmniGen2} \\
DreamOmni2~\citep{xia2025dreamomni2} & \texttt{\footnotesize xiabs/DreamOmni2} \\
BAGEL~\citep{deng2025bagel} & \texttt{\footnotesize ByteDance-Seed/BAGEL-7B-MoT} \\
FLUX.2 [klein] 4B~\citep{blackforestlabs2026flux2klein} & \texttt{\footnotesize black-forest-labs/FLUX.2-klein-4B} \\
FLUX.2 [klein] 9B~\citep{blackforestlabs2026flux2klein} & \texttt{\footnotesize black-forest-labs/FLUX.2-klein-9B} \\
Qwen-Image-Edit-2511~\citep{wu2025qwenimagetechnicalreport} & \texttt{\footnotesize Qwen/Qwen-Image-Edit-2511} \\
DyRef~\citep{huang2026scaling} & \texttt{\footnotesize Weistrass/Qwen-Image-Edit-2511-DyRef} \\
\midrule
\multicolumn{2}{l}{\textit{Proprietary image models}} \\
GPT-Image-1.5~\citep{openai2025gptimage1_5} & \texttt{\footnotesize gpt-image-1.5} \\
Nano Banana Pro~\citep{google2025nanobananapro} & \texttt{\footnotesize gemini-3-pro-image} \\
Seedream 4.5~\citep{bytedance2025seedream45} & \texttt{\footnotesize seedream-4-5-251128} \\
\midrule
\multicolumn{2}{l}{\textit{Reasoning models}} \\
GPT-5.5~\citep{openai2026gpt55} & \texttt{\footnotesize gpt-5.5-2026-04-23} \\
Qwen3-VL-32B-Instruct~\citep{bai2025qwen3vl} & \texttt{\footnotesize Qwen/Qwen3-VL-32B-Instruct} \\
\midrule
\multicolumn{2}{l}{\textit{Evaluator (MultiBanana)}} \\
Qwen3-VL-8B-Instruct~\citep{bai2025qwen3vl} & \texttt{\footnotesize Qwen/Qwen3-VL-8B-Instruct} \\
\midrule
\multicolumn{2}{l}{\textit{Evaluator (OmniContext)}} \\
GPT-4.1 & \texttt{\footnotesize gpt-4.1} \\
\midrule
\multicolumn{2}{l}{\textit{Proposer}} \\
Claude Fable 5.1 via Claude Code~\citep{anthropic2025claudecode} & \texttt{\footnotesize claude-fable-5-1} \\
\bottomrule
\end{tabular}
\end{table}

\section{Extended Related Work}
\label{app:related_tts}

\paragraph{Test-Time Scaling for Multimodal Generation.}
Test-time scaling (TTS), which improves model capabilities by allocating additional computation at inference time, has its roots in the development of reasoning in large language models~\citep{kojima2022large, snell2024scaling}. 
This paradigm has recently been extended to image and video generation, where a growing body of work improves generation quality and alignment with human preferences~\citep{furuta2024improving, onoda2026multiaxis} by scaling inference-time computation without updating model parameters~\citep{yeh2024sampling, kim2025das, ma2025inferencetime, singhal2025general, oshima2025inference, zhang2026classical, he2026evosearch}. Beyond uniformly increasing computation for all inputs, adaptive TTS dynamically allocates computation budgets according to input difficulty or intermediate evaluations, aiming to achieve more efficient search~\citep{wu2025imagerysearch, zhao2026latsearch, guimaraes2026psp, rawal2026flashbon, saini2026cachedsearch, jung2026inferencetime}, with applications spanning image editing~\citep{qu2026scalespeed} and joint audio-video generation~\citep{jung2026inferencetime}. 
From a broader perspective, self-improving agents and agentic refinement, which iteratively alternate between generation, evaluation, and revision, can also be viewed as a form of TTS that leverages additional inference-time computation to improve solutions~\citep{hao2023promptist, wang2024genartist, yao2026photoagent}. In this view, our work does not merely increase the search performed for each instance; it automatically optimizes the agent harness to find a more effective and computationally efficient inference procedure.

\paragraph{Image Generation and Editing Agents.}
Image generation agents combine prompt adaptation~\citep{hao2023promptist,datta2024prompt_expansion}, tool orchestration~\citep{shen2023hugginggpt,wang2024genartist,guo2025comfymind}, and visual feedback~\citep{yang2025idea2img,wan2025maestro,kovalev2025craft} to improve model outputs.
GEMS~\citep{he2026gems} integrates iterative generation with trajectory memory and reusable skills.
Related systems use tree search for multi-step editing~\citep{yao2026photoagent} or combine tool use and self-refinement with multi-reference composition~\citep{meta2026muse}.
Beyond runtime refinement, reusable agent policies can be learned through reinforcement learning or experience distillation~\citep{jiang2026genagent,chen2026genevolve}, with some approaches jointly training reasoning, tool use, and native image generation~\citep{zhao2026toolartist}.
Unlike these human-written or trained agents, AutoRef searches over the harness code itself while keeping all models frozen.

\section{The AutoRef Algorithm}
\label{app:algorithm}

\begin{algorithm}[H]
\caption{AutoRef: Harness Optimization for Multi-Reference Image Generation}
\label{alg:meta_harness}
\begin{algorithmic}[1]

\REQUIRE Frozen reasoning model $M_{\theta}$, image generator $G_{\phi}$, evaluator $R$, proposer $P$
\REQUIRE Training set $D_{\mathrm{train}}$, validation set $D_{\mathrm{val}}$, iterations $T$, beam width $B$, candidates per iteration $K$
\ENSURE Final beam $\mathcal{B}_T$ and search history $\mathcal{L}_T$

\STATE $\mathcal{B}_0 \leftarrow \{H_{\mathrm{vanilla}}, H_{\mathrm{init}}\}$

\STATE Run each $H \in \mathcal{B}_0$ on $D_{\mathrm{train}}$ and collect training evidence
\STATE Initialize $\mathcal{L}_0$ with the harnesses and their training evidence

\FOR{$t=0,\ldots,T-1$}

    \STATE $P$ inspects $\mathcal{B}_t$ and $\mathcal{L}_t$ and proposes
    $\mathcal{C}_{t+1} = \{H_{t+1}^{(k)}\}_{k=1}^{K}$
    \hfill {\color{richlilac} $\triangleright$ Propose candidate harnesses}
    \label{line:propose}

    \FOR{$k=1,\ldots,K$}
        \STATE Run $H_{t+1}^{(k)}$ on $D_{\mathrm{train}}$ and collect evidence $\mathcal{E}_{t+1}^{(k)}$
        \hfill {\color{richlilac} $\triangleright$ Collect proposer-visible training evidence}
        
        \label{line:train-evidence}
    \ENDFOR

    \FOR{$k=1,\ldots,K$}
        \STATE Evaluate $J_{D_{\mathrm{val}}}(H_{t+1}^{(k)})$
        \label{line:dev-eval}
    \ENDFOR

    \STATE $\mathcal{B}_{t+1}
        \leftarrow
        \operatorname{TopB}_{H \in \mathcal{C}_{t+1}}
        J_{D_{\mathrm{val}}}(H)$
        \hfill {\color{richlilac} $\triangleright$ Select on the hidden validation set}
        \label{line:beam-select}

    \STATE Append all candidate harnesses, training evidence, and selected candidate identities to $\mathcal{L}_{t+1}$; do not expose validation scores or artifacts
    \label{line:update-history}

\ENDFOR

\RETURN $(\mathcal{B}_T,\mathcal{L}_T)$

\end{algorithmic}
\end{algorithm}

Algorithm~\ref{alg:meta_harness} gives the full search loop of Section~\ref{sec:method}. 
The search starts with the base generator $H_{\mathrm{vanilla}}$ and the initial harness $H_{\mathrm{init}}$, both run on $D_{\mathrm{train}}$, so the first history already contains their outputs (lines 1--3). 
At each iteration, the proposer reads the current beam and the history and writes $K$ new harnesses as code (line~\ref{line:propose}). 
Each candidate is run on $D_{\mathrm{train}}$ (line~\ref{line:train-evidence}), and its code, per-task scores, evaluator rationales, execution trajectories, and generated images are added to the search history for later iterations (line~\ref{line:update-history}).
Each candidate is then scored on $D_{\mathrm{val}}$ (line~\ref{line:dev-eval}), and the $B$
candidates with the highest $J_{D_{\mathrm{val}}}$ form the next beam (line~\ref{line:beam-select}).
The history records which candidates were kept, but not their validation scores or outputs
(line~\ref{line:update-history}): the proposer learns which directions survived without seeing the
data that decided it. With $T=5$, $K=4$, and $B=2$, the search evaluates 20 candidates, and
AutoRef-Harness is the member of $\mathcal{B}_T$ with the highest $J_{D_{\mathrm{val}}}$.

\clearpage
\section{How the Harness Changed Across Iterations}
\label{app:lineage}

\begin{figure}[h]
\centering
\includegraphics[width=\textwidth]{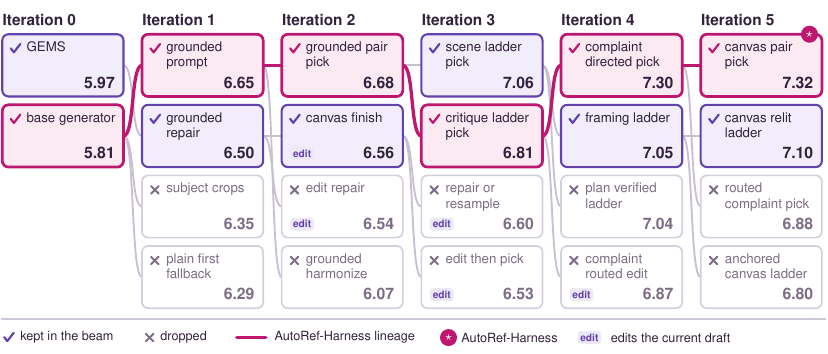}
\caption{\textbf{The AutoRef search that produced AutoRef-Harness.} The two initial harnesses and
the 20 candidates written over five iterations, each with its $J_{D_{\mathrm{val}}}$ score; each candidate has an
edge to its parent. 
Checked boxes were kept in the beam; the pink path is the lineage that ends in AutoRef-Harness ($\star$), and \emph{edit} marks candidates that repair a draft by editing it. 
Candidate names are those given by the proposer.}
\label{fig:autoref-lineage}
\end{figure}

\autoref{fig:autoref-lineage} shows the complete search: two initial harnesses (the base generator
and GEMS) and five iterations of $K=4$ candidates with a beam of $B=2$.

\paragraph{How AutoRef-Harness was assembled.}
The lineage of AutoRef-Harness (the pink path) acquired the components of
Section~\ref{sec:optimized_harness} over the five iterations; scores in \autoref{fig:autoref-lineage} are $J_{D_{\mathrm{val}}}$.
Iteration~1 replaced the raw instruction with a prompt written by the reasoning model that states
what each reference contributes, still with one image per task (reference-grounded prompting,
\S\ref{sec:optimized_grounding}; 5.81 $\rightarrow$ 6.65). Iteration~2 drew two drafts from two
differently framed grounded prompts and chose between them by a pairwise comparison run in both
presentation orders (6.68). Iteration~3 drew both drafts from a scene-first prompt and added a
third draft generated from a prompt revised for the current winner's weakest criterion (6.81);
the budget has remained at three images per task since. Iteration~4 based this revision on concrete
complaints checked against the references and applied a hard-failure check to every draft before
the pairwise comparison (complaint-directed revision and failure-aware selection,
\S\ref{sec:optimized_complaint}--\ref{sec:optimized_selection}; 7.30). Iteration~5 restored
structural diversity: one of the two drafts became canvas-anchored, with the reference that sets
the background or style passed to the generator first (structurally diverse drafts,
\S\ref{sec:optimized_generation}; 7.32).

\paragraph{Directions that were not retained.}
The \emph{edit} candidates repair the current draft by passing it to the generator as the first
image, followed by the references and an instruction to change only the failing element. This
direction entered at iteration~2 (edit repair, 6.54; canvas finish, 6.56, which also re-renders
the draft under a single light). Its descendants reached 6.53 and 6.60, and the direction left the beam at
iteration~3, and a later edit-based variant on the main lineage (complaint routed edit, 6.87) was
also dropped. The proposer's own analyses on the training tasks identify the causes: edits left
the flagged failure in place on 8 of 12 flagged tasks, and the single-light re-render lowered
the score of the draft it was applied to (6.16 $\rightarrow$ 5.91). Other single-step variants were dropped
at once: cropping small subjects from their references (subject crops, 6.35), generating from the
raw instruction and grounding only after a failure (plain first fallback, 6.29), and re-rendering
the grounded draft under one light (grounded harmonize, 6.07). A second lineage branched off at
iteration~3 (scene ladder pick, 7.06, three drafts from one scene-first prompt) and then wrote the
scene as a structured plan from which several framings of the prompt were rendered (framing ladder,
7.05; plan verified ladder, 7.04). It remained in the
beam until the end and, at iteration~5, also arrived at canvas-anchored drafts (canvas relit
ladder, 7.10), but did not reach AutoRef-Harness.

\begin{figure}[h]
\centering
\includegraphics[width=0.45\linewidth]{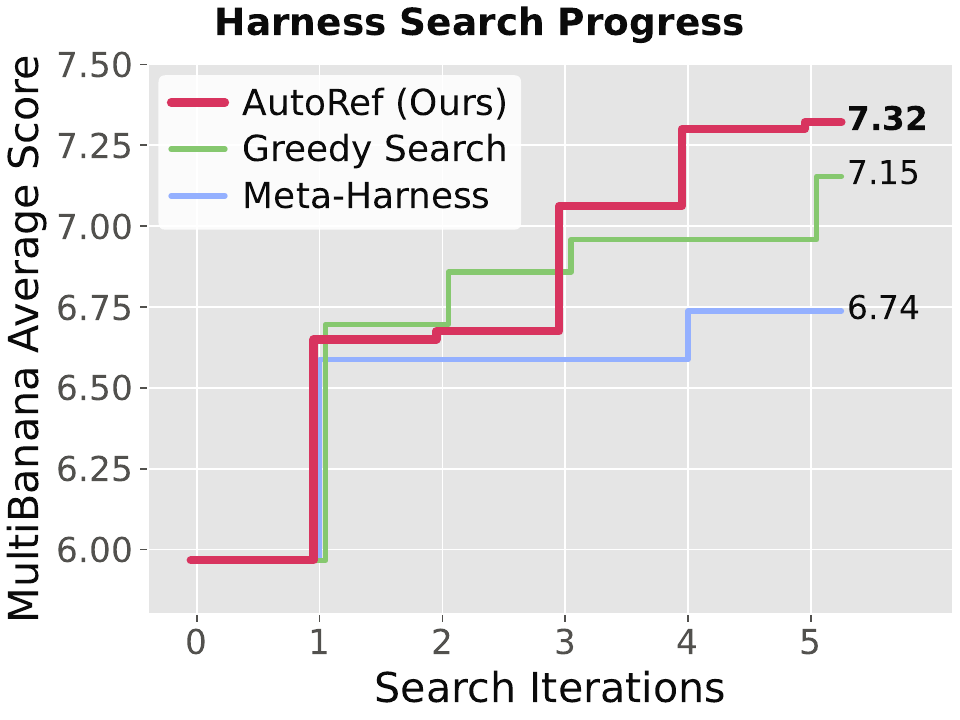}
\caption{\textbf{Search progress on $D_{\mathrm{val}}$.} Best validation score found so far at
each iteration by AutoRef, Greedy Search, and Meta-Harness, with the same proposer, models, and
evaluator. Meta-Harness does not select on $D_{\mathrm{val}}$; its candidates are scored on it only for
this figure.}
\label{fig:search-progress}
\end{figure}

\paragraph{Comparison with other search procedures.}
\autoref{fig:search-progress} shows the best $J_{D_{\mathrm{val}}}$ reached at each iteration by
AutoRef and by the two search baselines of Appendix~\ref{app:baselines_harness_search}. After five
iterations, AutoRef reaches 7.32, Greedy Search 7.15, and Meta-Harness 6.74.

\section{Prompts for Proposer Coding Agent}
\label{app:prompts}

The proposer is a coding agent that starts a new session, with no memory of earlier sessions, at every iteration. 
It receives two prompts: a system prompt that is fixed across iterations and provided as a Claude Code~\citep{anthropic2025claudecode} skill, and an iteration prompt that specifies the iteration, the number of tasks each candidate is evaluated on, the current beam, and the log files it may read. 
It gets everything else by reading files. 
In the excerpts below, split names are replaced by our notation.

\paragraph{Information available to the proposer.} The proposer runs in an isolated container. It can
read the code, scores, evaluator rationales, execution trajectories, and generated images of every earlier candidate on
$D_{\mathrm{train}}$, and the names of the candidates kept in the beam. Evaluation results on
$D_{\mathrm{val}}$ and on the held-out test split are stored outside the container and are never
exposed.

\paragraph{System prompt.} Most of the system prompt describes the task and the harness interface and
is adapted from the proposer skill of Meta-Harness~\citep{lee2026metaharness}. We reproduce the parts
that shape the search: the two model calls available to a harness, and the rules that determine what
counts as a valid candidate. In the excerpt, a harness is a Python class whose \texttt{run} method
maps a task to an output image, \texttt{ctx.think} calls the reasoning model, \texttt{ctx.generate}
calls the image generator, and \texttt{mean\_generations} is the reported number of images per task.

\begin{tcolorbox}[breakable, title=Proposer System Prompt (excerpt)]
[... the task, and the harness interface ...]

\vspace{0.5em}

\texttt{ctx} gives you, and nothing else:\\
- \texttt{ctx.think(prompt, images=[...]) -> str} -- the frozen LLM (gpt-5.5). \texttt{images} are bytes or
  paths; \texttt{<image>} placeholders in the prompt take them in order.\\
- \texttt{ctx.generate(prompt, images=[...]) -> bytes} -- the frozen image model. \texttt{images} are bytes or
  paths the model conditions on. Call it as many times as your mechanism needs; every call is
  counted and reported as \texttt{mean\_generations}.\\
- \texttt{ctx.calls} -- the running count of every call this run has made.

\vspace{0.5em}

\textbf{Anti-parameter-tuning rules}

\vspace{0.5em}

The most common failure mode is creating harnesses that are just parameter variants of existing
ones. Check \texttt{evolution\_summary.jsonl} for what's been tried -- sweeps (how many rounds, how many
questions, how many references to pass) almost always regress or tie.

\vspace{0.5em}

\textbf{Good candidates change a fundamental mechanism:}\\
- A new verification design (e.g. graded checks instead of yes/no, or checks that compare against
  a specific reference)\\
- A new refinement architecture (e.g. separate what must be kept from what must change)\\
- A new control flow (e.g. let the first verification decide how many rounds to run)\\
- A new rule for what to return

\vspace{0.5em}

\textbf{Bad candidates just tune numbers.} If \texttt{run()} is identical to an existing harness except for
constants, it is a parameter variant. Rewrite with a genuinely novel mechanism.

\vspace{0.5em}

\textbf{Combining harnesses is valid.} Take the verification from A and the refinement from B.

\vspace{0.5em}

\textbf{Anti-overfitting rules}

\vspace{0.5em}

- \textbf{No task-specific hints.} Do not hardcode knowledge about particular prompts or subjects.\\
- \textbf{Never mention the benchmark name} in harness code, prompts, or comments.\\
- \textbf{General patterns are OK.} ``Check the subject before the background'' applies broadly.\\
- \texttt{ctx} is your only access to the models.
\end{tcolorbox}

\paragraph{Iteration prompt.} This is the only prompt that changes between iterations. It specifies
the iteration, the number of tasks per candidate, earlier iterations' logs,
the current beam, and the number of candidates per beam member. The example
below is from iteration~5; paths are placeholders.

\begin{tcolorbox}[breakable, title=Iteration Prompt (iteration 5)]

Run iteration 5 of the harness evolution loop for task `multibanana'. Each candidate will be benchmarked on 48 tasks of $D_{\mathrm{train}}$.

\vspace{0.5em}

\textbf{Run directories}\\
All logs and results for this run are under \texttt{<logs>/}.\\
- \texttt{<logs>/evolution\_summary.jsonl} - past results\\
- \texttt{<logs>/frontier.json} - frontier\\
- \texttt{<logs>/runs/<harness>.json/harness\_result.json} - per-harness scores,
  tool counts, worst tasks\\
- \texttt{<logs>/runs/<harness>.json/trace/<id>.json} - every tool call, per task\\
- \texttt{<logs>/reports/} - post-eval reports\\
- Harness files live in \texttt{<harnesses>/}\\
- Write \texttt{pending\_eval.json} to: \texttt{<logs>/pending\_eval.json}\\
- \texttt{<logs>/adoption.jsonl} - which candidates were kept as base harnesses,
  and which were not\\
- \texttt{<logs>/beam.json} - the current base harnesses

\vspace{0.5em}

\textbf{Base harnesses}\\
The current base harnesses are \texttt{complaint\_directed\_pick}, \texttt{framing\_ladder} (files in
\texttt{<harnesses>/}). Write exactly 4 candidates: 2 that build on each base. Each
candidate's \texttt{base\_harness} is the exact name of its base. Which candidates are kept as bases for
the next iteration is decided on tasks outside $D_{\mathrm{train}}$ that this session cannot see:
the 2 candidates that do best there replace the current bases, so a gain has to hold on unseen
tasks to be kept.
\end{tcolorbox}

\clearpage
\section{The Flow of AutoRef-Harness}
\label{app:harness-flow}

\begin{figure}[h]
\centering
\includegraphics[width=\textwidth]{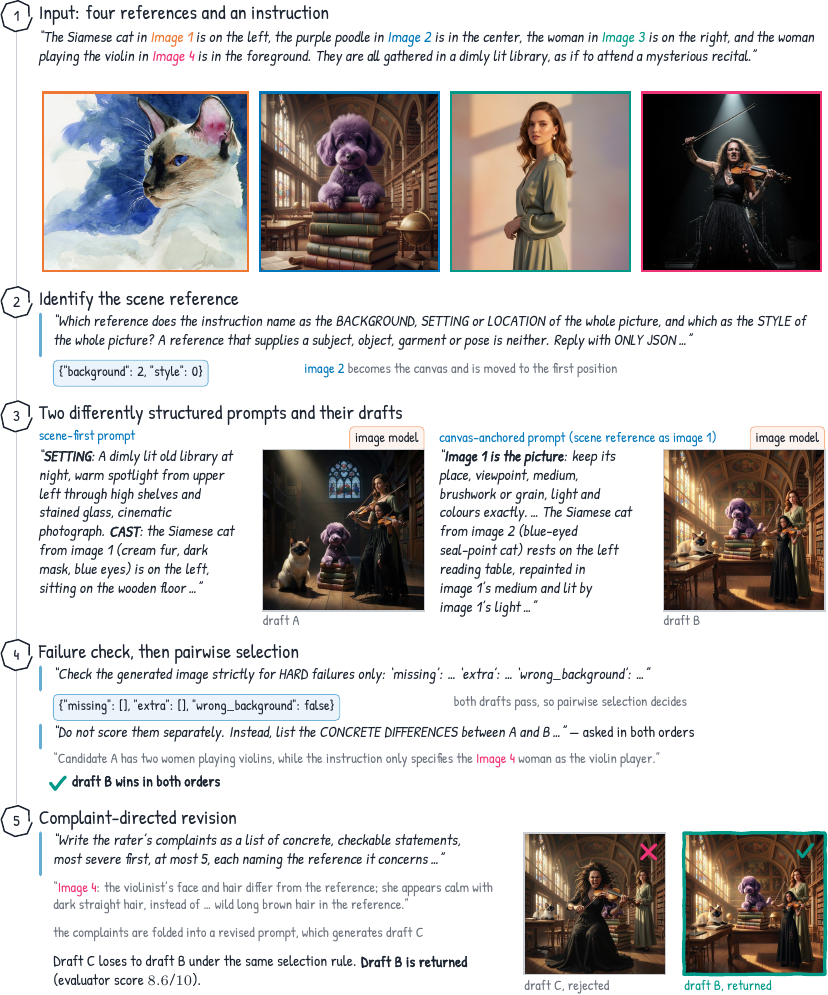}
\caption{\textbf{Execution trajectory of AutoRef-Harness on a held-out four-reference task.} Steps 1--5
show the input, the identification of the scene reference, the two drafts, failure-aware selection,
and complaint-directed revision. Quoted text with a blue bar is a prompt sent to the reasoning model
$M_\theta$, and the text below it is the model's response; boxes labeled ``image model'' mark calls to
the generator $G_\phi$. Prompts and responses are abridged but not otherwise modified. Each reference's
frame color matches that of its number in the text.}
\label{fig:harness-flow}
\end{figure}

\autoref{fig:harness-flow} shows how AutoRef-Harness processes one task. 
The reasoning model first infers that image~2, the poodle in a library, sets the scene (step~2). 
It then writes two structurally different prompts (step~3): a scene-first prompt that describes the setting and then each subject (draft~A), and a canvas-anchored prompt that passes image~2 first and places the other subjects in it (draft~B). 
Both drafts pass the failure check, and draft~B is preferred in both presentation orders (step~4). The reasoning model lists complaints about draft~B, mainly that the violinist in image~4 does not match her reference and is not in the foreground, and its revised prompt produces draft~C (step~5). Draft~C does not beat draft~B under the same rule, so draft~B is returned (evaluator score 8.6/10).


\section{Baselines}
\label{app:baselines}

\subsection{Human-Written Harness Baselines}
\label{app:baselines_human_written_harness}

All four human-written baselines use the same generator and reasoning model as AutoRef-Harness; adaptations to the multi-reference setting are noted per method.

\paragraph{Best-of-$N$~\citep{ma2025inferencetime}.} $N$ images are sampled independently from the
original instruction, and the reasoning model picks one in one call showing the
references and candidates, using a priority-ordered rubric (every requested
subject present, fidelity to each reference, correct background, consistent lighting, realism). We
use $N=3$ throughout, matching AutoRef-Harness's three images.

\paragraph{GEMS~\citep{he2026gems}.} An agentic loop with skills and memory. The instruction is routed
to a matching skill and decomposed into yes/no requirement questions; each round generates an image,
checks every question against it, stops if all pass, and otherwise summarizes the round into memory
and rewrites the prompt from the accumulated history. The image satisfying the most questions is
returned. We use the published prompts and the default four rounds, and show the verifier the
references beside the image, since the original loop verifies single-image text-to-image
outputs.

\paragraph{IPR~\citep{oshima2026multibanana}.} Iterative Prompt Refinement, the agentic baseline
proposed with MultiBanana. Over three steps, each step generates from the current prompt, and a
planner refines the prompt from the references and the image just generated:
\[
  y^{t+1} = \mathrm{Gen}(u^{t}, \mathcal{I}), \qquad
  u^{t+1} = \mathrm{Plan}(u^{t}, \mathcal{I}, y^{t+1}).
\]
The generator never sees earlier images, and the last image is returned.
No code was released, so we re-implement it from its equations and prompts, with
our reasoning model as the planner.

\paragraph{Idea2Img~\citep{yang2025idea2img}.} Iterative self-refinement in which a multimodal model
drafts several prompts, selects the best image, and writes feedback that, with a memory of
earlier prompts and feedback, guides the next round. We keep the official budget of three prompts over
three rounds with a final selection among round winners (nine images per task). As this triples
AutoRef-Harness's budget, we also report a budget-matched variant: three prompts, one round, no feedback.

\subsection{Harness Search Baselines}
\label{app:baselines_harness_search}

\paragraph{Meta-Harness~\citep{lee2026metaharness}.} Our implementation of the Meta-Harness protocol, run with the same proposer,
models, and evaluator as AutoRef. The proposer sees the full history of earlier
candidates --- their code, scores, execution trajectories, and images --- and the candidates are
ranked on the same tasks whose feedback the proposer reads, so there is no separate selection split.
It has no explicit parents: the proposer chooses which candidate
to build on, and the best candidate on these tasks is reported. We run it for five iterations, as for AutoRef (the original runs 20--40 iterations).

\paragraph{Greedy Search.} AutoRef with a beam of one. Proposal and selection use separate splits as
in AutoRef ($D_{\mathrm{train}}$ for feedback, $D_{\mathrm{val}}$ for selection), but only the single
best candidate on $D_{\mathrm{val}}$ is kept at each iteration and becomes the parent of all
candidates in the next iteration.

\clearpage

\setcounter{topnumber}{3}
\section{Further Results}
\label{sec:further_results}

\subsection{Detailed MultiBanana Results}
\label{sec:multibanana_per_metric}
The Qwen3-VL-8B-Instruct~\citep{bai2025qwen3vl} evaluator of MultiBanana scores each image on five
criteria (instruction alignment, reference consistency, background--subject match, physical realism,
and visual quality), and the main text reports their mean. Tables~\ref{tab:multibanana-4ref_appendix}--\ref{tab:multibanana-5ref_appendix} report each
criterion for four, three, and five references, averaged over task types; per-type averages are in
Tables~\ref{tab:multibanana-summary-4ref} and~\ref{tab:multibanana-summary-3-5ref}.

\begin{table}[ht]
\centering
\caption{MultiBanana with 4 references, held-out test split ($n{=}133$), by evaluation metric. Each metric is averaged over the four task types (object, local, global, background), and Avg.\ is the mean of the five metrics, as in the Avg.\ column of Tables~\ref{tab:multibanana-summary-4ref} and~\ref{tab:multibanana-summary-3-5ref}. \emph{Gen.}\ is images drawn per task. Inst.: instruction alignment; Ref.: reference consistency; BG: background--subject match; Real.: physical realism; Qual.: visual quality. Best open image generators per column in bold, second best underlined.}
\label{tab:multibanana-4ref_appendix}
\small
\begin{tabular}{l c cccccc}
\toprule
Method & Gen. & Inst. & Ref. & BG & Real. & Qual. & Avg. \\
\midrule
GPT-Image-1.5 & 1 & 6.58 & 7.63 & 6.49 & 6.79 & 7.88 & 7.07 \\
Nano Banana Pro & 1 & 6.94 & 7.74 & 6.62 & 6.78 & 7.94 & 7.20 \\
Seedream 4.5 & 1 & 6.62 & 7.78 & 6.38 & 6.73 & 7.66 & 7.03 \\
\midrule
OmniGen2 & 1 & 3.08 & 3.36 & 3.16 & 3.87 & 5.26 & 3.75 \\
DreamOmni2 & 1 & 3.36 & 3.57 & 3.10 & 3.91 & 5.22 & 3.83 \\
BAGEL & 1 & 2.83 & 3.13 & 2.66 & 3.22 & 3.92 & 3.15 \\
\midrule
FLUX.2 [klein] 4B & 1 & 5.25 & 5.86 & 5.04 & 5.64 & 6.83 & 5.72 \\
\quad $+$ Best-of-3~\citep{ma2025inferencetime} & 3 & 5.37 & 6.08 & 5.50 & 6.08 & 7.02 & 6.01 \\
\quad $+$ GEMS~\citep{he2026gems} & 2.7 & 4.92 & 5.43 & 4.74 & 5.38 & 6.39 & 5.37 \\
\quad $+$ IPR~\citep{oshima2026multibanana} & 3 & 6.42 & 7.08 & 6.67 & 7.10 & 7.84 & 7.02 \\
\quad $+$ Idea2Img~\citep{yang2025idea2img} & 9 & \underline{6.50} & \textbf{7.39} & 6.71 & 7.00 & 7.98 & \underline{7.12} \\
\quad $+$ Idea2Img, budget-matched & 3 & 5.89 & 6.85 & 5.90 & 6.24 & 7.20 & 6.42 \\
\quad $+$ Meta-Harness~\citep{lee2026metaharness} & 4.2 & 5.64 & 6.22 & 5.84 & 6.29 & 7.18 & 6.23 \\
\quad $+$ Greedy Search & 5 & 5.90 & 6.82 & 6.31 & 6.72 & 7.58 & 6.67 \\
\quad $+$ AutoRef (2nd) & 3 & 6.15 & 6.75 & \underline{6.88} & 7.37 & 7.85 & 7.00 \\
\quad $+$ \textbf{AutoRef-Harness} & 3 & \textbf{6.57} & \underline{7.15} & \textbf{7.20} & \textbf{7.77} & \textbf{8.16} & \textbf{7.37} \\
\quad $+$ \textbf{AutoRef-Harness}, Qwen3-VL-32B & 3 & 6.22 & 6.96 & 6.46 & 7.14 & 7.70 & 6.90 \\
\midrule
FLUX.2 [klein] 9B & 1 & 5.19 & 5.96 & 5.28 & 5.80 & 6.70 & 5.78 \\
\quad $+$ \textbf{AutoRef-Harness} & 3 & 6.48 & 6.85 & 6.71 & \underline{7.46} & \underline{7.99} & 7.10 \\
\midrule
Qwen-Image-Edit-2511 & 1 & 3.95 & 4.63 & 3.66 & 4.28 & 5.69 & 4.44 \\
\quad $+$ \textbf{AutoRef-Harness} & 3 & 5.11 & 5.78 & 5.15 & 5.82 & 6.59 & 5.69 \\
\quad $+$ DyRef~\citep{huang2026scaling} & 1 & 4.77 & 4.89 & 4.23 & 4.81 & 5.86 & 4.91 \\
\quad $+$ DyRef $+$ \textbf{AutoRef-Harness} & 3 & 5.68 & 5.66 & 5.22 & 6.12 & 6.83 & 5.90 \\
\bottomrule
\end{tabular}
\end{table}

\begin{table}[ht]
\centering
\caption{MultiBanana with 3 references ($n{=}96$), by evaluation metric. Columns as in \autoref{tab:multibanana-4ref_appendix}.}
\label{tab:multibanana-3ref_appendix}
\small
\begin{tabular}{l c cccccc}
\toprule
Method & Gen. & Inst. & Ref. & BG & Real. & Qual. & Avg. \\
\midrule
GPT-Image-1.5 & 1 & 7.22 & 8.31 & 7.56 & 7.84 & 8.35 & 7.86 \\
Nano Banana Pro & 1 & 6.79 & 7.91 & 7.15 & 7.68 & 7.99 & 7.50 \\
Seedream 4.5 & 1 & 6.95 & 8.47 & 6.58 & 7.06 & 8.00 & 7.41 \\
\midrule
OmniGen2 & 1 & 4.55 & 5.15 & 4.51 & 5.46 & 6.58 & 5.25 \\
DreamOmni2 & 1 & 4.75 & 5.20 & 4.67 & 5.56 & 6.64 & 5.36 \\
BAGEL & 1 & 3.77 & 4.14 & 3.36 & 4.16 & 5.05 & 4.10 \\
\midrule
FLUX.2 [klein] 4B & 1 & 6.03 & 6.96 & 6.59 & 7.31 & 7.82 & 6.94 \\
\quad $+$ Best-of-3 & 3 & 6.14 & 7.46 & 6.51 & 6.88 & 7.80 & 6.96 \\
\quad $+$ \textbf{AutoRef-Harness} & 3 & \underline{6.99} & \underline{7.69} & \underline{7.74} & \underline{8.03} & 8.38 & \underline{7.76} \\
\quad $+$ \textbf{AutoRef-Harness}, Qwen3-VL-32B & 3 & 6.71 & \textbf{7.71} & 7.27 & 7.84 & \underline{8.41} & 7.59 \\
\midrule
FLUX.2 [klein] 9B & 1 & 6.18 & 7.27 & 6.31 & 6.74 & 7.63 & 6.83 \\
\quad $+$ \textbf{AutoRef-Harness} & 3 & \textbf{7.05} & 7.66 & \textbf{7.75} & \textbf{8.21} & \textbf{8.45} & \textbf{7.82} \\
\midrule
Qwen-Image-Edit-2511 & 1 & 4.53 & 5.48 & 4.10 & 4.95 & 6.25 & 5.06 \\
\quad $+$ \textbf{AutoRef-Harness} & 3 & 6.25 & 7.16 & 6.48 & 7.05 & 7.79 & 6.95 \\
\bottomrule
\end{tabular}
\end{table}

\begin{table}[ht]
\centering
\caption{MultiBanana with 5 references ($n{=}96$), by evaluation metric. Columns as in \autoref{tab:multibanana-4ref_appendix}.}
\label{tab:multibanana-5ref_appendix}
\small
\begin{tabular}{l c cccccc}
\toprule
Method & Gen. & Inst. & Ref. & BG & Real. & Qual. & Avg. \\
\midrule
GPT-Image-1.5 & 1 & 6.73 & 7.45 & 5.65 & 5.96 & 7.20 & 6.60 \\
Nano Banana Pro & 1 & 6.82 & 7.50 & 5.66 & 6.21 & 7.36 & 6.71 \\
Seedream 4.5 & 1 & 6.64 & 7.36 & 5.80 & 6.02 & 7.19 & 6.60 \\
\midrule
OmniGen2 & 1 & 2.84 & 2.66 & 2.59 & 3.20 & 4.60 & 3.18 \\
DreamOmni2 & 1 & 2.54 & 3.10 & 2.70 & 3.17 & 4.08 & 3.12 \\
BAGEL & 1 & 3.00 & 2.73 & 2.17 & 2.61 & 3.57 & 2.82 \\
\midrule
FLUX.2 [klein] 4B & 1 & 5.01 & 5.22 & 4.38 & 4.94 & 6.43 & 5.19 \\
\quad $+$ Best-of-3 & 3 & 5.42 & 5.39 & 4.54 & 5.19 & 6.40 & 5.39 \\
\quad $+$ \textbf{AutoRef-Harness} & 3 & \underline{6.35} & \underline{6.24} & \underline{5.55} & \textbf{6.29} & \textbf{7.36} & \underline{6.36} \\
\quad $+$ \textbf{AutoRef-Harness}, Qwen3-VL-32B & 3 & 5.72 & 6.14 & 5.07 & 5.81 & 6.79 & 5.91 \\
\midrule
FLUX.2 [klein] 9B & 1 & 5.65 & 5.79 & 4.72 & 5.31 & 6.56 & 5.61 \\
\quad $+$ \textbf{AutoRef-Harness} & 3 & \textbf{6.59} & \textbf{6.91} & \textbf{5.59} & \underline{6.08} & \underline{7.16} & \textbf{6.47} \\
\midrule
Qwen-Image-Edit-2511 & 1 & 1.74 & 2.04 & 1.55 & 1.85 & 2.24 & 1.89 \\
\quad $+$ \textbf{AutoRef-Harness} & 3 & 1.72 & 2.02 & 1.45 & 1.70 & 2.47 & 1.87 \\
\bottomrule
\end{tabular}
\end{table}

\subsection{Component Ablation by Task Type}
\autoref{tab:ablation-axes} gives the per-type scores behind \autoref{tab:ablation}. Removing any
one of the three components lowers the score on every task type, and without reference-grounded
prompting (selection only) the score falls below all three leave-one-out rows on every task type.

\begin{table}[h]
\centering
\caption{Component ablation on the held-out test split, by task type. Rows as in \autoref{tab:ablation};
Avg.\ is the mean over the four task types, as reported there. All rows draw three images per task except Grounding only and Generator only, which draw one.}
\label{tab:ablation-axes}
\begin{tabular}{l ccccc}
\toprule
 & Object & Local & Global & Background & Avg. \\
\midrule
\textbf{Full harness} & 7.27 & 7.87 & 7.64 & 6.70 & \textbf{7.37} \\
\midrule
\quad $-$ diverse drafts & 7.07 & 7.37 & 7.14 & 6.47 & 7.01 \\
\quad $-$ complaint revision & 7.24 & 7.01 & 7.33 & 6.38 & 6.99 \\
\quad $-$ failure-aware selection & 6.79 & 7.39 & 7.55 & 6.00 & 6.93 \\
\midrule
Grounding only & 7.21 & 6.72 & 7.21 & 6.50 & 6.91 \\
Selection only & 6.11 & 6.78 & 6.14 & 5.56 & 6.15 \\
Generator only & 5.95 & 6.11 & 5.50 & 5.34 & 5.72 \\
\bottomrule
\end{tabular}
\end{table}

\subsection{Compatibility with Fine-Tuning}
\label{app:finetune}

\begin{figure}[ht]
    \centering
    \includegraphics[width=0.45\linewidth]{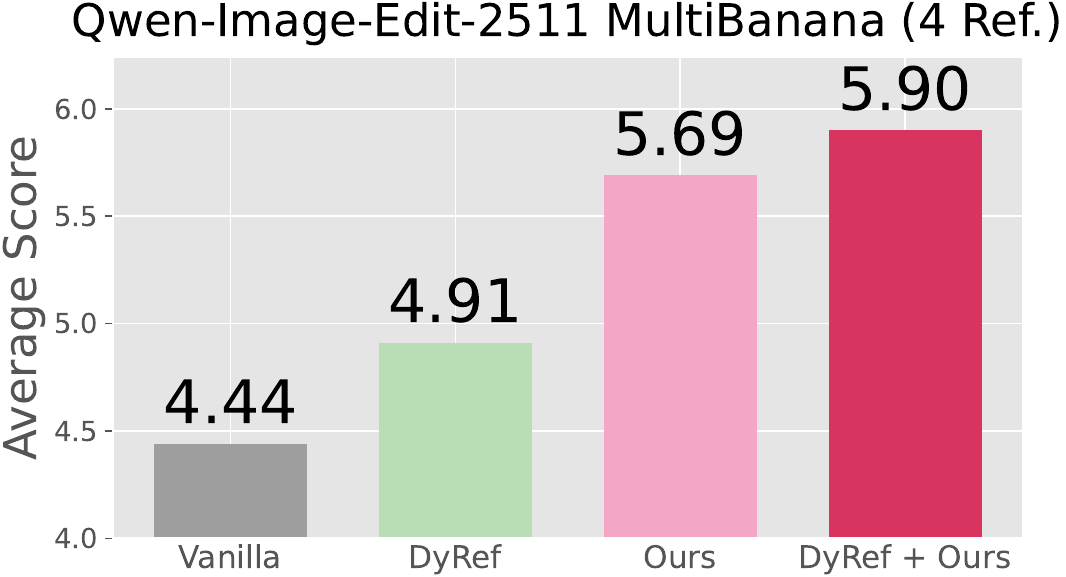}
    \caption{
    \textbf{Harness optimization is compatible with fine-tuning.}
    On Qwen-Image-Edit-2511, AutoRef-Harness outperforms DyRef alone, and applying it to the DyRef-trained model yields a further improvement. Ours denotes AutoRef-Harness.
    }
    \label{fig:bar_comparison_dyref}
\end{figure}

We ask whether harness optimization remains useful when the underlying image generator is already optimized for multi-reference image generation.
DyRef~\citep{huang2026scaling} improves Qwen-Image-Edit-2511 through supervised fine-tuning followed by reward optimization.
As shown in \autoref{fig:bar_comparison_dyref}, AutoRef-Harness applied to the original Qwen-Image-Edit-2511 achieves 5.69, outperforming DyRef alone at 4.91.
Applying the same harness to the DyRef-trained model further improves performance to 5.90.
These results suggest that harness optimization and model-weight optimization provide complementary gains and can be combined.

\subsection{Detailed OmniContext Results}
Tables~\ref{tab:omnicontext_appendix} and~\ref{tab:omnicontext-scene_appendix} provide the full OmniContext breakdown behind \autoref{tab:omnicontext-overall}, reporting prompt following (PF), subject consistency (SC), and their geometric mean for each task type.

\begin{table*}[ht]
\centering
\caption{OmniContext, SINGLE and MULTIPLE task types, 15 tasks each: prompt following, subject consistency, and their geometric mean behind \autoref{tab:omnicontext-overall}. \emph{Gen.}\ is images drawn per task. Best open image generators per column in bold, second best underlined.}
\label{tab:omnicontext_appendix}
\resizebox{\textwidth}{!}{%
\begin{tabular}{l c ccc ccc ccc ccc ccc }
\toprule
 & & \multicolumn{6}{c}{SINGLE} & \multicolumn{9}{c}{MULTIPLE} \\
\cmidrule(lr){3-8} \cmidrule(lr){9-17}
 & & \multicolumn{3}{c}{Character} & \multicolumn{3}{c}{Object} & \multicolumn{3}{c}{Character} & \multicolumn{3}{c}{Object} & \multicolumn{3}{c}{Char. + Obj.} \\
\cmidrule(lr){3-5}\cmidrule(lr){6-8}\cmidrule(lr){9-11}\cmidrule(lr){12-14}\cmidrule(lr){15-17}
Method & Gen. & PF & SC & Overall & PF & SC & Overall & PF & SC & Overall & PF & SC & Overall & PF & SC & Overall \\
\midrule
GPT-Image-1.5 & 1 & 9.80 & 9.33 & 9.56 & 9.80 & 9.60 & 9.70 & 9.67 & 9.00 & 9.32 & 9.67 & 9.27 & 9.46 & 9.27 & 9.27 & 9.26 \\
Nano Banana Pro & 1 & 9.53 & 9.73 & 9.63 & 9.53 & 9.33 & 9.42 & 9.73 & 9.20 & 9.46 & 9.40 & 9.00 & 9.19 & 9.00 & 9.07 & 9.02 \\
Seedream 4.5 & 1 & 9.60 & 9.33 & 9.45 & 9.47 & 9.60 & 9.50 & 9.20 & 9.13 & 9.09 & 9.67 & 9.13 & 9.39 & 9.13 & 9.07 & 9.09 \\
\midrule
OmniGen2 & 1 & 8.20 & 8.93 & 8.51 & 6.87 & 6.40 & 5.73 & 7.13 & 5.87 & 6.30 & 7.67 & 5.87 & 6.58 & 7.80 & 7.67 & 7.71 \\
DreamOmni2 & 1 & 7.53 & 8.33 & 7.81 & 7.20 & 6.60 & 6.72 & 5.07 & 5.13 & 4.80 & 7.00 & 7.40 & 7.07 & 6.73 & 5.53 & 5.92 \\
BAGEL & 1 & 8.20 & 6.40 & 6.67 & 6.67 & 8.53 & 7.09 & 4.27 & 3.07 & 3.43 & 7.07 & 6.80 & 6.74 & 7.20 & 7.20 & 7.08 \\
\midrule
FLUX.2 [klein] 4B & 1 & \underline{9.40} & \underline{9.07} & \underline{9.22} & 8.27 & 8.87 & 8.16 & 8.00 & 7.93 & 7.91 & 8.87 & 7.73 & 8.21 & 8.47 & \textbf{8.93} & 8.68 \\
\quad $+$ Best-of-3 & 3 & \underline{9.40} & \textbf{9.13} & \textbf{9.26} & 8.60 & 8.20 & 7.97 & 8.73 & 8.73 & 8.71 & 8.87 & \underline{8.87} & 8.81 & 8.73 & \underline{8.87} & \underline{8.79} \\
\quad $+$ \textbf{AutoRef-Harness} & 3 & 9.33 & 8.60 & 8.94 & 9.20 & \textbf{9.00} & 8.99 & \underline{9.33} & 8.73 & 9.02 & 9.00 & 8.60 & 8.78 & 8.53 & 8.67 & 8.59 \\
\midrule
FLUX.2 [klein] 9B & 1 & \underline{9.40} & 9.00 & 9.17 & 9.33 & \underline{8.93} & 9.08 & 9.00 & 8.40 & 8.66 & 8.73 & 7.73 & 8.14 & \textbf{8.87} & 8.67 & 8.74 \\
\quad $+$ \textbf{AutoRef-Harness} & 3 & \textbf{9.47} & 9.00 & \underline{9.22} & \underline{9.53} & 8.87 & \underline{9.18} & \textbf{9.40} & \textbf{9.00} & \textbf{9.19} & \textbf{9.73} & \textbf{8.93} & \textbf{9.32} & \underline{8.80} & \underline{8.87} & \textbf{8.82} \\
\midrule
Qwen-Image-Edit-2511 & 1 & \underline{9.40} & 8.93 & 9.14 & \textbf{9.73} & 8.73 & \textbf{9.19} & 8.87 & 8.53 & 8.66 & \underline{9.60} & 8.47 & \underline{9.00} & 8.20 & 8.53 & 8.33 \\
\quad $+$ \textbf{AutoRef-Harness} & 3 & 9.27 & 9.00 & 9.12 & 9.13 & 8.47 & 8.64 & \underline{9.33} & \underline{8.87} & \underline{9.09} & 9.13 & 8.20 & 8.62 & 8.47 & 8.27 & 8.34 \\
\bottomrule
\end{tabular}}
\end{table*}

\begin{table*}[ht]
\centering
\caption{OmniContext, SCENE task types. Columns as in \autoref{tab:omnicontext_appendix}.}
\label{tab:omnicontext-scene_appendix}
\resizebox{\textwidth}{!}{%
\begin{tabular}{l c ccc ccc ccc }
\toprule
 & & \multicolumn{9}{c}{SCENE} \\
\cmidrule(lr){3-11}
 & & \multicolumn{3}{c}{Character} & \multicolumn{3}{c}{Object} & \multicolumn{3}{c}{Char. + Obj.} \\
\cmidrule(lr){3-5}\cmidrule(lr){6-8}\cmidrule(lr){9-11}
Method & Gen. & PF & SC & Overall & PF & SC & Overall & PF & SC & Overall \\
\midrule
GPT-Image-1.5 & 1 & 10.00 & 9.40 & 9.69 & 9.53 & 9.27 & 9.39 & 8.80 & 9.13 & 8.93 \\
Nano Banana Pro & 1 & 9.73 & 9.00 & 9.35 & 8.07 & 8.87 & 8.39 & 7.93 & 8.53 & 8.20 \\
Seedream 4.5 & 1 & 9.87 & 8.87 & 9.35 & 8.60 & 8.80 & 8.66 & 8.13 & 8.47 & 8.23 \\
\midrule
OmniGen2 & 1 & 7.27 & 6.80 & 6.93 & 6.47 & 5.87 & 6.10 & 7.53 & 6.73 & 7.03 \\
DreamOmni2 & 1 & 6.40 & 5.40 & 5.78 & 6.40 & 5.07 & 5.63 & 6.20 & 5.40 & 5.72 \\
BAGEL & 1 & 4.87 & 4.00 & 3.97 & 4.13 & 4.20 & 4.11 & 5.73 & 5.67 & 5.63 \\
\midrule
FLUX.2 [klein] 4B & 1 & 9.67 & 8.73 & 9.18 & 7.40 & 7.67 & 7.47 & 7.40 & 7.73 & 7.54 \\
\quad $+$ Best-of-3 & 3 & \underline{9.73} & \textbf{8.93} & \underline{9.32} & 8.73 & 8.33 & 8.49 & 7.93 & 7.80 & 7.74 \\
\quad $+$ \textbf{AutoRef-Harness} & 3 & \underline{9.73} & \underline{8.87} & 9.28 & \underline{9.00} & \textbf{8.80} & \textbf{8.89} & 8.27 & \textbf{8.33} & \underline{8.28} \\
\midrule
FLUX.2 [klein] 9B & 1 & \textbf{9.87} & \underline{8.87} & \textbf{9.35} & 7.87 & \underline{8.47} & 8.07 & 7.27 & 7.80 & 7.47 \\
\quad $+$ \textbf{AutoRef-Harness} & 3 & \textbf{9.87} & \underline{8.87} & \textbf{9.35} & 8.80 & 8.07 & 8.40 & \textbf{8.60} & \underline{8.20} & \textbf{8.30} \\
\midrule
Qwen-Image-Edit-2511 & 1 & 7.67 & 6.60 & 6.97 & 8.40 & 8.07 & 8.17 & \underline{8.53} & 7.87 & 8.17 \\
\quad $+$ \textbf{AutoRef-Harness} & 3 & 9.07 & 8.40 & 8.71 & \textbf{9.27} & 8.40 & \underline{8.82} & 8.40 & \underline{8.20} & \underline{8.28} \\
\bottomrule
\end{tabular}}
\end{table*}

\clearpage
\setcounter{topnumber}{2}

\section{Where the Harness Cannot Help}
\label{app:breakdown}

\begin{figure}[h]
\centering
\includegraphics[width=\textwidth]{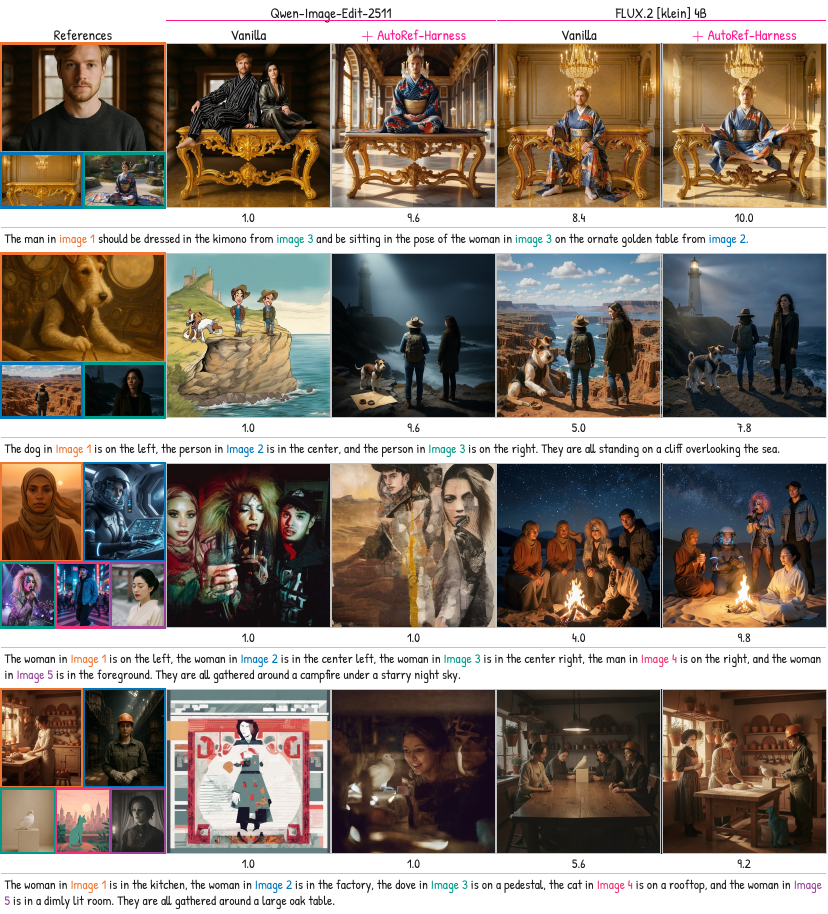}
\caption{\textbf{AutoRef-Harness helps Qwen-Image-Edit-2511 at three references but not at five.} Outputs of Qwen-Image-Edit-2511 and FLUX.2 [klein] 4B, each without and with AutoRef-Harness, on three-reference (top two rows) and five-reference (bottom two rows) tasks. The number below each output is its evaluator score (1--10).}
\label{fig:qwen-breakdown}
\end{figure}

AutoRef-Harness does not change the generator; it can only return one of the images the frozen
generator produces. Qwen-Image-Edit-2511 exposes this limit (\autoref{fig:qwen-breakdown}): the harness
raises its score from 5.06 to 6.95 at three references and from 4.44 to 5.69 at four,
but not at five (from 1.89 to 1.87; Tables~\ref{tab:multibanana-summary-4ref}
and~\ref{tab:multibanana-summary-3-5ref}). All harness steps still run at five references, but the
generator rarely produces an acceptable image: the hard-failure check (\S\ref{sec:optimized_selection}) flags all three drafts on 87 of 96
tasks (91\%), versus 9 of 96 (9\%) for FLUX.2 [klein] 4B at five references and 7 of
96 (7\%) for Qwen-Image-Edit-2511 at three. When no candidate is acceptable, better selection
cannot help; overcoming this limit likely requires reducing how many references the generator
must compose at once.

\clearpage
\section{Further Qualitative Comparisons}
\label{app:qualitative}

\begin{figure}[p]
\centering
\includegraphics[width=\textwidth]{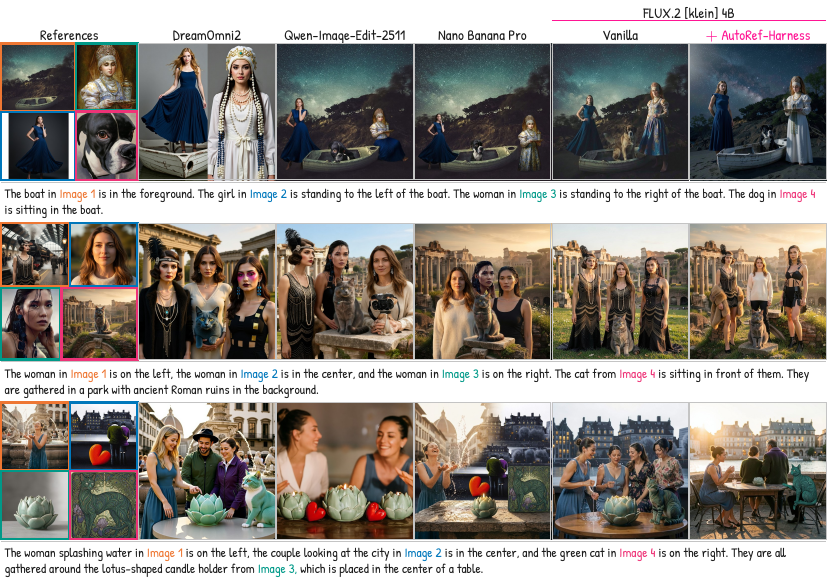}
\caption{\textbf{Object composition.} Held-out four-reference tasks that compose all subjects into one scene.}
\label{fig:qual-object}
\end{figure}

\begin{figure}[p]
\centering
\includegraphics[width=\textwidth]{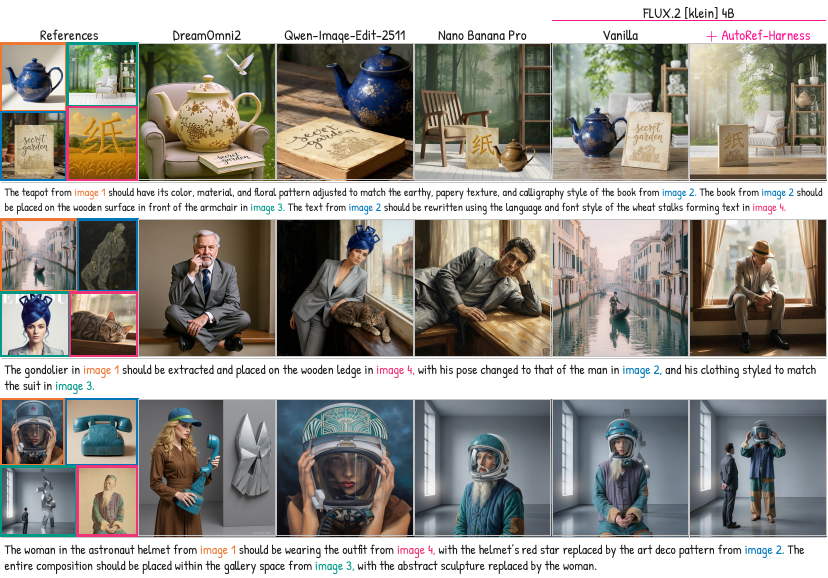}
\caption{\textbf{Local attribute transfer.} Held-out four-reference tasks in which an attribute of one reference (e.g., a pose, a garment, a texture, or a text style) is applied to another.}
\label{fig:qual-local}
\end{figure}

\begin{figure}[p]
\centering
\includegraphics[width=\textwidth]{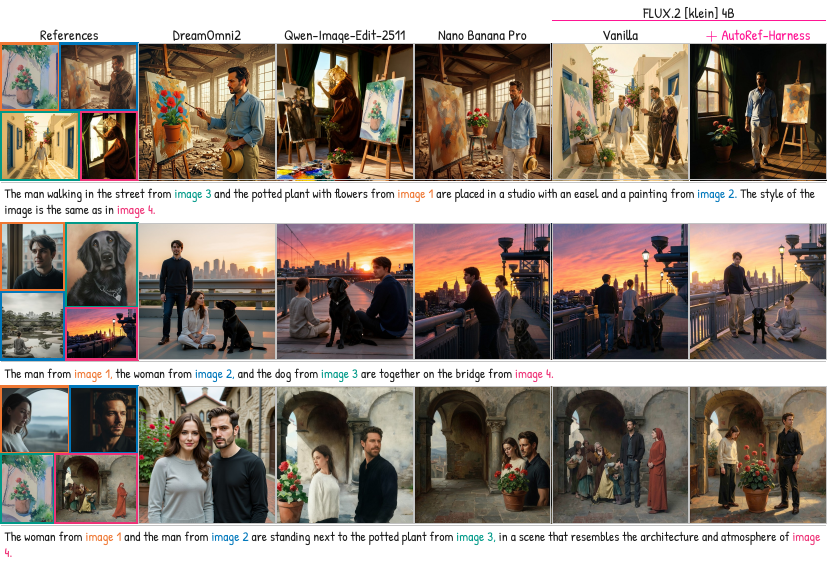}
\caption{\textbf{Background and global style.} Held-out four-reference tasks in which one reference sets the background or the style of the whole image.}
\label{fig:qual-scene}
\end{figure}

\begin{figure}[p]
\centering
\includegraphics[width=\textwidth]{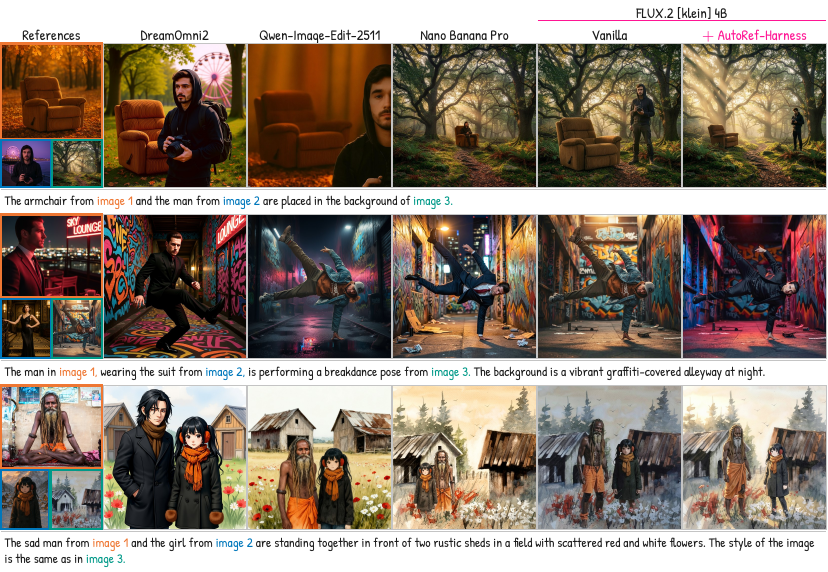}
\caption{\textbf{Three references.} MultiBanana tasks with three references, a count not used during the search.}
\label{fig:qual-ref3}
\end{figure}

\begin{figure}[p]
\centering
\includegraphics[width=\textwidth]{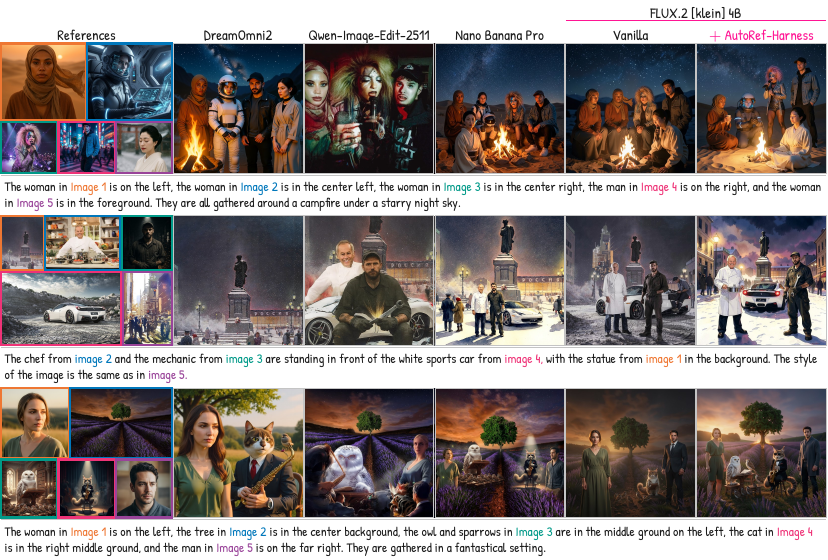}
\caption{\textbf{Five references.} MultiBanana tasks with five references, a count not used during the search.}
\label{fig:qual-ref5}
\end{figure}

\begin{figure}[p]
\centering
\includegraphics[width=\textwidth]{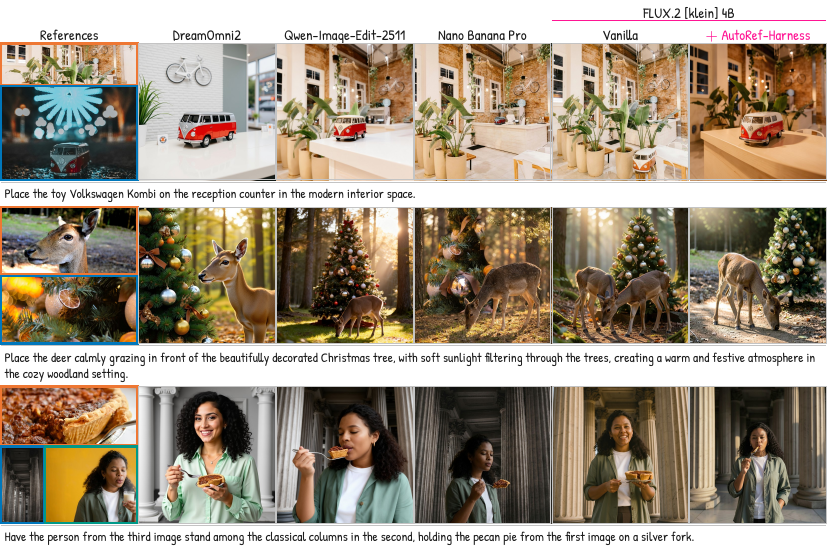}
\caption{\textbf{OmniContext.} Tasks with two or three references, from a benchmark not used during the search.}
\label{fig:qual-omni}
\end{figure}

All figures in this section follow the layout of \autoref{fig:qualitative}: the references on the
left, each framed in a distinct color that also marks its number in the instruction (when the instruction numbers the references), the outputs of five methods,
and the full instruction below. Figures~\ref{fig:qual-object}--\ref{fig:qual-scene} show three held-out four-reference tasks each, for object composition,
local attribute transfer, and background and global style; Figures~\ref{fig:qual-ref3} and~\ref{fig:qual-ref5} show tasks with three and five references, counts not used during the search;
and \autoref{fig:qual-omni} shows OmniContext. None of these tasks was seen during
the search. They were selected among the tasks with the largest score gap between the base generator and
AutoRef-Harness, so they illustrate the failures the harness removes and are not a random sample;
aggregate results are in \autoref{sec:further_results}. In all settings, the harness removes the
same kinds of failure: a requested reference is dropped or used in the wrong role.

\end{document}